\documentclass[letterpaper, 10 pt, conference]{ieeeconf}

\IEEEoverridecommandlockouts                              

\usepackage{caption}
\usepackage{xcolor}
\usepackage{cite}
\usepackage{graphicx}
\usepackage{amsmath}
\usepackage{textcomp}
\usepackage{algorithm}
\usepackage{amssymb}
\usepackage{textcomp}
\usepackage{algpseudocode}
\usepackage{tabularx} 
\usepackage{soul,xcolor}
\usepackage[nolist]{acronym}
\usepackage{booktabs}
\usepackage{tabularx}
\usepackage{array}

\makeatletter
\let\NAT@parse\undefined
\makeatother
\usepackage[hidelinks]{hyperref}
\hypersetup{
    colorlinks=true,
    linkcolor=blue,
    citecolor=blue,
    filecolor=magenta,      
    urlcolor=cyan,
    pdftitle={Overleaf Example},
    pdfpagemode=FullScreen,
    }

\title{\LARGE \bf 
Mapping the Wind to Find the Source: Online Physics-Informed Gas Source Localization with Mobile Robots }
\author{Wanting Jin, Thomas Wiedemann, Han Fan, Achim Lilienthal, Alcherio Martinoli
}
\begin{acronym}
    \acro{CFD}{Computational Fluid Dynamic}
    \acro{GMRF}{Gaussian Markov Random Field}
    \acro{GSL}{Gas Source Localization}
    \acro{GDM}{Gas Distribution Mapping}
    \acro{IPP}{Informative Path Planning}
    \acro{MAE}{Mean Absolute Error}
    \acro{MCS}{Motion Capture System}
    \acro{MOX}{Metal OXide}
    \acro{PDEs}{Partial Differential Equations}
    \acro{PDE}{Partial Differential Equation}
    \acro{PID}{Photo-Ionization Detector}
    \acro{PCB}{Printed Circuit Board}
    \acro{RMSE}{Root Mean Square Error}
    \acro{ROS}{Robot Operating System}
    \acro{STE}{Source Term Estimation}
    \acro{ADE}{Advection-Diffusion Equation}
    \acro{DDPM}{Data-Driven Plume Model}
    \acro{FEM}{Finite Element Method}
    \acro{FVM}{Finite Volume Method}
    \acro{FDM}{Finite Difference Method}
    \acro{GKR}{Gaussian Kernel Regression}
    \acro{PDF}{Probability Density Function}
    \acro{DBSCAN}{Density-Based Spatial Clustering of Applications with Noise}
    \acro{NS}{Navier-Stokes}
    \acro{MAP}{Maximum a Posteriori}
    \acro{GP}{Gaussian Process}
    \acro{GPM}{Gaussian Plume Model}
    \acro{FBM}{Filament-based Model}
    \acro{SLAM}{Simultaneous Localization and Mapping}
\end{acronym}
\begin{document}
\bstctlcite{IEEEexample:BSTcontrol}
\setstcolor{red}
\maketitle
\thispagestyle{empty}
\pagestyle{empty}

\begin{abstract}
\ac{GSL} with mobile robots is challenging because wind and gas measurements are spatially sparse, noisy, and potentially affected by robot motion. Indoor airflow can vary substantially across space, while detailed environmental layouts, boundary conditions, or precomputed airflow fields are often unavailable. Moreover, hazardous sources must be localized as quickly as possible, making exhaustive exploration impractical. This paper presents a lightweight physics-informed probabilistic framework for efficient \ac{GSL} under spatially varying indoor airflow. As the robot navigates, it continuously collects wind and gas measurements and reconstructs the spatially varying airflow field online. The estimated wind field drives a forward dispersion model that predicts the gas pattern for each candidate source location. These predictions are compared with the measured gas pattern to update a probabilistic belief over the source location. As the wind field is reconstructed incrementally, the framework also accounts for the spatially non-uniform reliability of wind field estimation and its influence on the source estimation. The resulting belief map is then used for informative path planning to guide the robot toward informative regions. High-fidelity simulations provide a comprehensive evaluation of the individual components and their integrations within the complete system, while physical indoor experiments assess the system performance under realistic sensing, motion, and airflow conditions. The results demonstrate that incorporating lightweight physical knowledge significantly improves localization robustness under spatially complex airflow and enables more efficient exploration.
\end{abstract}

\section{Introduction}
Gas leaks pose serious risks to human health, infrastructure, and the environment~\cite{jing_recent_2021}. Mobile robots equipped with gas sensors can reduce human exposure by autonomously exploring potentially hazardous areas. \ac{GSL} refers to the task of inferring the spatial location of a gas source from a limited set of gas measurements. As \ac{GSL} is often time-critical, exhaustively sampling the entire environment is impractical. Instead, the robot should use the available measurements efficiently and prioritize informative regions. Since gas transport is governed by physical processes, incorporating knowledge of gas dispersion into source inference can constrain the search and guide the robot navigation more efficiently, particularly during the early stages of an emergency when measurements are limited.


Probabilistic \ac{GSL} methods estimate a spatial belief over possible source locations from the measurements collected by robots \cite{jain_review_2025}. In model-based approaches, a gas dispersion model is incorporated into the source estimation process to relate candidate source hypotheses to the observed gas measurements.
Gas dispersion is governed primarily by airflow and turbulent mixing: the wind field determines the dominant transport direction and strongly influences the shape of the resulting plume~\cite{rossi_gas_2017}. Consequently, constructing a gas dispersion model under spatially complex airflow requires an estimate of the wind field. 

Constructing wind maps and estimating gas dispersion models online is particularly challenging in indoor environments. Walls, obstacles, and ventilation systems can produce airflow that varies substantially across space, violating the spatially uniform airflow assumptions adopted by many analytical plume models \cite{jin_towards_2023}. High fidelity \ac{CFD} techniques can represent such conditions, but generally require detailed boundary conditions, extensive environmental knowledge, and considerable computational resources. Furthermore, increasing model fidelity does not necessarily improve source localization in practice. More sophisticated models generally require additional environmental parameters and boundary information, whose uncertainty can introduce further model mismatch. This challenge is compounded by wind and gas measurements that are noisy, collected incrementally, and affected by robot motion.

\begin{figure} [t]
    \centering
    \includegraphics[width=\linewidth]{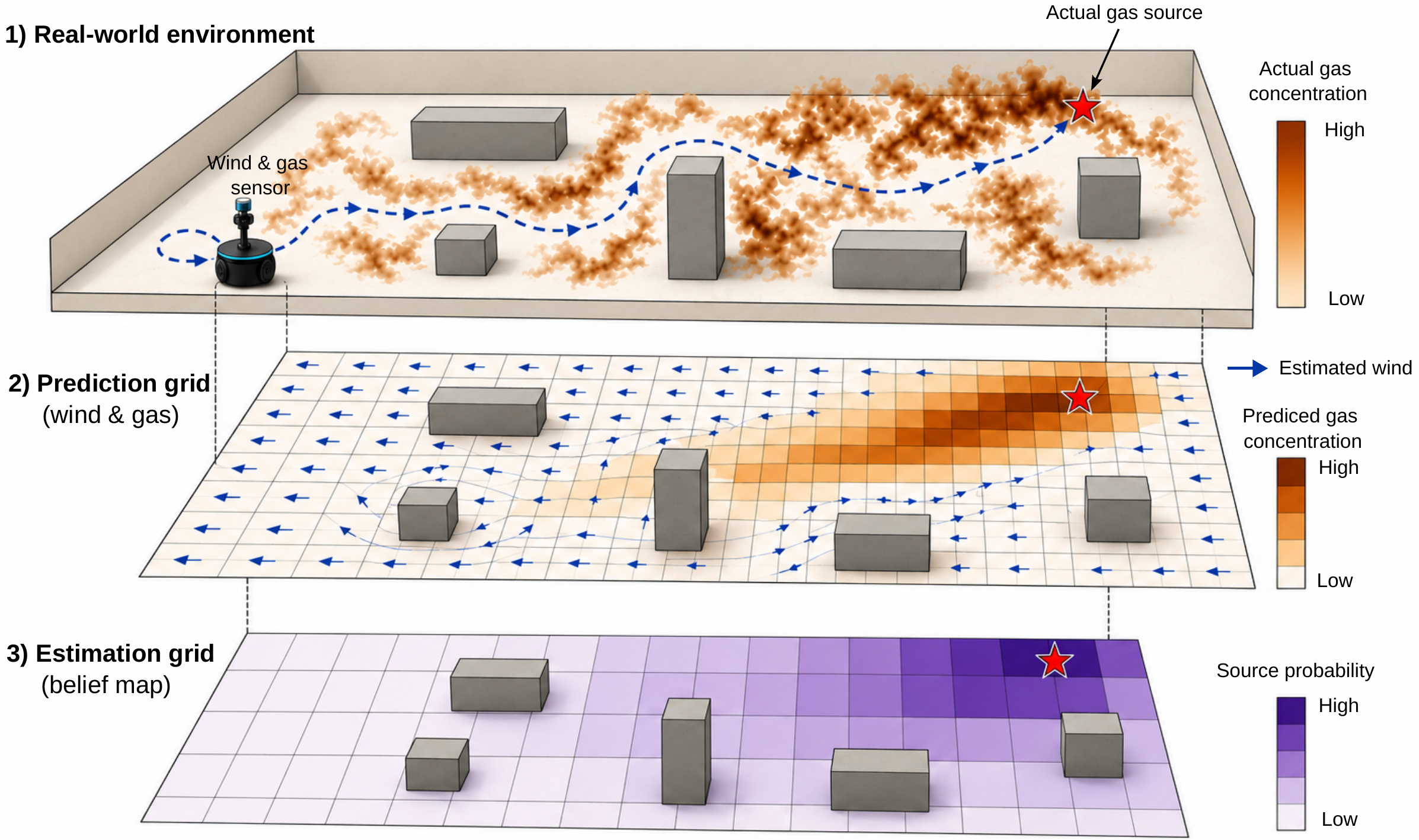}
    \caption{Overview of the proposed physics-informed \ac{GSL} framework. A fine prediction grid supports wind field reconstruction and gas dispersion prediction, while a coarser estimation grid represents the candidate source locations and their posterior probabilities.}
    \label{fig:pipeline_overview}
\end{figure}
 
Rather than reconstructing the complete wind and gas concentration fields with high fidelity, as commonly pursued in gas mapping \cite{ercolani_gaslam_2022} or wind field mapping tasks \cite{francis_gas_2022}, this work investigates a task-oriented perspective: can a lightweight physics-based model provide sufficient information for online \ac{GSL} with limited environmental knowledge? An additional challenge arises because the wind field is reconstructed incrementally. During the early stages of exploration, the wind estimate is generally more reliable near the robot’s trajectory than in unvisited regions. This spatially non-uniform reliability of the wind estimation also affects the predicted gas distributions and, consequently, the inferred source belief. Ignoring it can produce an overconfident posterior that prematurely suppresses the true source hypothesis and directs the robot away from informative regions.


To address these challenges, we propose a physics-informed probabilistic GSL framework illustrated in Fig.~\ref{fig:pipeline_overview}. A mobile robot equipped with an anemometer and an in-situ gas sensor continuously collects wind and gas measurements while navigating through the environment. The wind measurements are used to reconstruct the airflow field online. The airflow field then drives an \ac{ADE}-based forward model to predict the resulting gas distribution originating from each candidate source location, yielding a source-conditioned prediction. These predictions are compared with the measured gas pattern to update a probabilistic belief map over the source location. To account for the spatially non-uniform reliability of the reconstructed wind field, we derive a wind-support-aware uncertainty, which is measured from the wind mapping process and we use it to adjust the influence of each source-conditioned prediction on the likelihood. The resulting belief can subsequently guide the robot toward informative regions.

This work considers a two-dimensional indoor environment containing a single stationary gas source and assumes that the time-averaged airflow and gas distribution remain approximately stationary during a localization task. The formal assumptions are stated in Sec.~\ref{sec:problem_statement}. The main contributions of this work are:
\begin{itemize}
    \item A lightweight online pipeline that couples kernel-based wind field reconstruction with an ADE-based forward dispersion model, enabling spatially varying airflow to be incorporated into probabilistic GSL without requiring detailed inlet, outlet, or flow boundary information.
    \item 
    A wind-support-aware likelihood formulation that accounts for the spatially non-uniform wind estimation confidence and its impact on downstream gas dispersion prediction and source location estimation.
    \item An integrated online \ac{GSL} system that incorporates wind and gas measurements collected continuously during robot motion and uses the estimated belief map to guide \ac{IPP}.
     \item A systematic evaluation in high-fidelity simulations and physical indoor experiments, including ablation studies of the dispersion model, online wind estimation, wind-support-aware inference, and their integration with \ac{IPP}.
\end{itemize}

The remainder of this paper is organized as follows. Sec.~\ref{sec: related_work} reviews probabilistic \ac{GSL} methods and existing approaches on wind field reconstruction and gas dispersion modeling. Sec.~\ref{sec:problem_statement} formulates the localization problem and states the underlying assumptions. Sec.~\ref{sec:wind_gas_modeling} introduces the online wind field reconstruction, and \ac{ADE}-based gas dispersion prediction. Sec.~\ref{sec:probabilistic_gsl} presents the relative gas feature representation, Bayesian source inference, and wind-support-aware likelihood formulation. Sec.~\ref{sec:continuous_sensing_planning} then describes continuous sensing, \ac{IPP}, and the termination conditions. Secs.~\ref{sec:simulation_exp} and~\ref{sec:physical_exp} report the simulation and physical experimental evaluations, respectively. Finally, Sec.~\ref{sec:conclusion} summarizes the main findings, discusses the limitations, and outlines directions for future work.

\section{Related Work} \label{sec: related_work}
This section reviews how physical knowledge is incorporated into robotic \ac{GSL}, with particular emphasis on the two components required by model-based probabilistic methods: reconstructing the wind field and predicting gas dispersion.

\subsection{Physics-Aware and Probabilistic Gas Source Localization}
Reactive methods are largely explored in the early \ac{GSL} work.  Bio-inspired \cite{rahbar20173}\cite{neumann_micro-drone_2011} and anemotaxis-based strategies \cite{gong_localising_2012}\cite{ishida_odor-source_1996} use local gas detections and wind measurements to move the robot in the upwind direction to trace back the gas source. 
These approaches are computationally lightweight and require little prior environmental knowledge. However, their reliance on local cues can lead to failure in cluttered environments, where obstacles and recirculating airflow may create local traps or misleading upwind directions \cite{hernandez_bennetts_mobile_2012}.

Probabilistic methods instead estimate a belief distribution over candidate source locations and update it as measurements are collected. By explicitly
representing uncertainty and integrating information over time, these methods
can be more robust to sensing noise than purely reactive strategies. 
Physical knowledge can be incorporated into this inference process at either a local or global level. In \cite{ojeda_information-driven_2021}, when the robot detects the gas, local wind measurements are used to propagate source belief in the upwind direction while accounting for obstacles. Although this method avoids constructing a global gas dispersion model, its dependence on local airflow information can be problematic under spatially complex wind conditions~\cite{ojeda_robotic_2023}.

On the other hand, model-based \ac{GSL} uses more global physics knowledge and formulates source localization as an inverse problem: source parameters are inferred from observations through a model that maps a
source hypothesis to the resulting gas distribution. Broadly, this inverse problem can be addressed in two ways. In direct inverse estimation, the source parameters are estimated by solving a regularized optimization or Bayesian identification problem constrained by the governing dispersion model. For example, \cite{wiedemann_model-based_2019} formulates GSL as sparse Bayesian identification of a diffusion \ac{PDE}, jointly estimating source number, locations, and release strengths from concentration measurements. Such inverse problems can be ill-posed because sparse and noisy observations may be compatible with multiple source configurations.

An alternative is forward-model-based probabilistic inference. Here, the predicted observations are generated for each candidate source hypothesis with the forward dispersion model, and the resulting predictions are compared with the observations to construct a likelihood over source locations. Infotaxis~\cite{vergassola_infotaxis_2007} uses a stochastic gas transport model to provide the expected encounter rate for a hypothesized source location, and the history of gas detections is used to update a Bayesian posterior over the source position. Similarly, \ac{STE} evaluates candidate source hypotheses by comparing measured gas concentrations with predictions generated by analytical dispersion models \cite{hutchinson2017review, rahbar2019algorithm, park_source_2022, bourne_decentralized_2020}. These methods are efficient under their assumed conditions, but they commonly rely on spatially uniform airflow. Some outdoor approaches \cite{li_odor_2011}\cite{Hinsen_2023} allow the wind to vary over time while approximating it as spatially homogeneous at each instant. This assumption is less suitable indoors, where walls, obstacles, and ventilation systems can produce strong spatial variations in both wind direction and magnitude.

Several methods have therefore incorporated environmental influences
more explicitly. A naïve physics approach was introduced in \cite{kowadlo_using_2006} to approximate airflow around obstacles. The \ac{DDPM} proposed in \cite{jin_towards_2023} uses a neural network trained on \ac{CFD} generated data to predict gas distributions from a source hypothesis and an obstacle layout in a wind tunnel environment. The efficiency of \ac{DDPM} is evaluated with the \ac{STE} framework under high fidelity simulations and physical experiments in a wind tunnel. It supports fast online prediction, but its generalization is constrained by the range of geometries and airflow conditions represented in the training data. In \cite{ojeda_robotic_2023}, \ac{GMRF} is used to reconstruct the wind field online, and a filament model simulates gas dispersion for candidate source locations. The resulting gas hit predictions are compared with binary measurements to infer the source position. More recently, \cite{tian_learning_2025} propose an end-to-end model that directly estimates the source location distribution from gas measurements, wind measurements, and the environmental layout.

These studies demonstrate the benefit of incorporating spatial airflow and gas dispersion information into probabilistic \ac{GSL}. However, model-based approaches must balance physical fidelity, online computational cost, and the limited information available during exploration. 


\subsection{Online Wind Field Reconstruction}
Wind field reconstruction methods can be broadly divided into physics-based and statistical approaches. Physics-based methods incorporate governing equations or physical constraints into the estimation process. In \cite{wiedemann_physics-informed_2025}, airflow reconstruction is formulated as a Bayesian maximum a posteriori inverse problem constrained by the steady incompressible \ac{NS} equations. Sparse wind measurements gathered by robots are fused with a finite element flow model, and the estimated uncertainty is used to guide robot exploration. The \ac{GMRF} formulation in \cite{monroy_online_2017} combines anemometer measurements with smoothness, incompressibility, and obstacle constraints and is subsequently used for online gas dispersion simulation in \cite{ojeda_robotic_2023}. More recently, \cite{monroy_physics-informed_2026} additionally incorporates an advection-energy term in the \ac{GMRF} formulation based on the estimated upstream/downstream direction. Alternatively, mesh-free radial basis functions have also been used to reconstruct incompressible ocean current fields \cite{jmse13050841}. These approaches can produce physically consistent estimates, but they may require knowledge of obstacle geometry, boundary conditions, or inlet and outlet locations. Their performance can also depend on regularization parameters that balance measurement agreement against different physical constraints.

Statistical approaches infer spatial correlations directly from measurements without solving the governing flow equations. \ac{GP}, for example, provide both field predictions and associated uncertainty at unobserved locations \cite{ghaffari2018gaussian}. However, exact \ac{GP} inference scales cubically with the number of observations, which can be restrictive for high frequency online sensing unless sparse or local approximations are introduced.

Kernel-based methods provide a lightweight alternative. Kernel DM+V and its extensions use local Gaussian kernels to propagate measurements to neighboring locations and estimate spatial mean and variance maps. Kernel-based methods have been widely used for gas distribution mapping tasks, where scattered gas measurements are interpolated to construct spatial gas distribution maps~\cite{lilienthal2009statistical}\cite{ercolani_gaslam_2022}. Because each new observation affects only a local neighborhood, these methods support incremental updates and scale well with both the number of measurements and the map resolution. Kernel geometry can also incorporate physical information. For example, \cite{reggente20103d}, uses anisotropic kernels aligned with the wind direction when modeling three-dimensional gas distributions.

Nevertheless, any wind field reconstructed incrementally from sparse measurements has spatially varying reliability: estimates are generally better supported near the robot’s trajectory than in unexplored regions. This issue is critical to our work as the uncertainty in the wind estimate also affects the reliability of source-conditioned gas predictions.

\subsection{Forward Gas Dispersion Models}
Gas transport is commonly modeled using the \ac{ADE}, a \ac{PDE} describing advective transport by the airflow together with diffusive mixing. In complex environments, the \ac{ADE} generally requires numerical solution. Analytical plume models provide computationally efficient solutions or approximations under restrictive assumptions. For example, the \ac{GPM} used in \cite{hutchinson2017review, rahbar2019algorithm, park_source_2022, bourne_decentralized_2020} and isotropic plume models used in \cite{vergassola_infotaxis_2007} \cite{hutchinson2019source} generally assume simplified airflow and homogeneous turbulence. Although well-suited for online execution, these models cannot directly represent spatially varying airflow or the influence of obstacles.

Filament-based models \cite{farrell_filament-based_2002} represent a plume as discrete gas parcels transported by the wind and diffused by small eddies. They reproduce intermittent plume structure more realistically and are widely used in robotic olfaction simulators \cite{Thomas2010}\cite{monroy_gaden_2017}. However, repeatedly simulating a filament plume for every candidate source can be computationally expensive. The method in \cite{ojeda_robotic_2023}, for example, employs a simplified two-dimensional binary filament model and a coarse-to-fine source search to reduce this cost.

Alternatively, the \ac{ADE} can be solved numerically using finite-difference~\cite{Hinsen_2023}\cite{wiedemann_model-based_2019}, finite-volume, or finite-element methods \cite{wiedemann_gas_2022}. Numerical solutions can incorporate spatially varying wind fields and obstacles while offering an adjustable trade-off between resolution, computational cost, and model fidelity. This makes them a promising intermediate representation between restrictive analytical plume models and computationally expensive filament or \ac{CFD} simulations.

\subsection{Research Gap}
There is still limited work combining online wind reconstruction and model-based probabilistic GSL, and the existing ones face two important limitations. First, many require prior environmental information, such as obstacle geometry, inlet/outlet locations, airflow boundary conditions, or a precomputed wind field. Such information may not be available beforehand. Second, when the wind field is reconstructed online from incrementally collected measurements, its reliability is spatially non-uniform: estimates are generally better supported near the robot’s trajectory than in unexplored regions. The uncertainty of the gas dispersion prediction caused by the wind estimation is typically ignored by the current existing work. 

This work addresses both limitations through a lightweight, physics-informed pipeline. The wind field is reconstructed locally from online wind measurements without requiring prior obstacle or airflow information. Geometric information acquired online through mapping techniques can optionally be incorporated into the wind estimation. The reconstructed wind field implicitly captures flow disturbances induced by the obstacles, allowing the ADE model to provide task-oriented dispersion predictions without explicitly specifying obstacle boundary conditions. In addition, the local measurement support for the wind estimate is incorporated into the likelihood uncertainty, reducing the influence of inaccurate predictions from poorly explored regions.

\section{Problem Statement and Framework Overview} \label{sec:problem_statement}
This section defines the problem and presents an overview of the proposed framework.
\subsection{Problem Statement}
Consider a bounded two-dimensional $L_x \times L_y$ indoor workspace $\Omega \subset \mathbb{R}^2$ containing a single stationary gas source at an unknown location $\textbf{x}_s$. Let  $\mathcal{O}$ denote the obstacles and  $\mathcal{F}=\Omega\setminus\mathcal{O}$ the free space. No prior knowledge of inlet/outlet conditions or global wind field is assumed. Obstacle geometry can be assumed unknown and obtained incrementally using external mapping techniques. In the experiments, we use a predefined occupancy map for collision-free navigation.

The source is assumed to be sufficiently far from the outer boundary of the search region and to have been active long enough for the time-averaged gas distribution to be approximated as steady state. The airflow is assumed to remain steady or vary sufficiently slowly over the duration of a localization task. The problem is formulated in two dimensions; consequently, vertical variations in airflow and gas concentration are not explicitly modeled.

The workspace is discretized using two grids with different resolutions and purposes: an \textbf{estimation grid} and a \textbf{prediction grid}. The coarse estimation grid containing $N_e$ cells in total, is defined as $\mathcal{G}^{\mathrm{est}} = \{E_1,...,E_{N_e} \}$, where $\mathbf{c}_i^{est}$ denotes the center of cell $E_i$. Each free-space cell represents a candidate source location. We define a discrete random variable $S\in \{1, ..., N_e\}$, where the event $S=i$ indicates that the gas source is located at $\mathbf{c}_i^{est}$. Cells known to
be occupied or located within a prescribed distance of the outer boundary are excluded from the source hypothesis space and assigned zero probability.

Gas and wind measurements are collected continuously as the robot moves through the free space. Let $\mathcal{T}_k$ denote the set of sample indices acquired during iteration $k$. The corresponding measurement batch is
\[
    \mathcal{Y}_k
    =
    \left\{
        \left(z^g_t,\mathbf{z}^w_t,\mathbf{r}_t\right)
    \right\}_{t\in\mathcal{T}_k},
    \label{eq:measurement_batch}
\]
where $z^g_t \in \mathbb{R}$ is the gas sensor measurement, $\mathbf{z}^w_t \in \mathbb{R}^{2}$ is the planar wind vector measurement, and $\mathbf{r}_t \in \mathcal{F}$ is the robot position at sample $t$. The measurement history available until iteration $k$ is denoted by $\mathcal{Y}_{1:k}=\{\mathcal{Y}_1,\ldots,\mathcal{Y}_k\}$. 
The finer prediction grid is denoted by $\mathcal{G}^{\mathrm{pred}}=\{P_1,\ldots,P_{N_p}\}$ where $N_p$ is the number of prediction cells and $\mathbf{c}^{\mathrm{pred}}_j$ denotes the center of cell $P_j$. This grid defines the spatial resolution at which the wind field and source-conditioned gas distributions are represented. 

The \ac{GSL} problem is to infer the posterior distribution over the candidate
source cells from the accumulated measurements. At iteration $k$, this
distribution is represented by the belief map
\begin{equation}
    b_k(i)
    =
    \Pr\!\left(S=i\mid\mathcal{Y}_{1:k}\right),
    \qquad i=1,\ldots,N_e.
    \label{eq:belief_map}
\end{equation}
At iteration $k$, the source location is estimated from the maximum a
posteriori source hypothesis:
\begin{equation}
    \hat{i}
    =
    \underset{i\in\{1,\ldots,N_e\}}{\arg\max}\; b_k(i),
    \qquad
    \hat{\mathbf{x}}_s
    =
    \mathbf{c}^{\mathrm{est}}_{\hat{i}}.
    \label{eq:map_source_estimate}
\end{equation}
The goal is to obtain a sufficiently confident source estimate within the limited iterations, according to the source location declaration criterion
defined in Sec.~\ref{subsec:termination}. 

\subsection{Framework Overview}
\begin{figure*}
    \centering
    \includegraphics[width=0.9\linewidth]{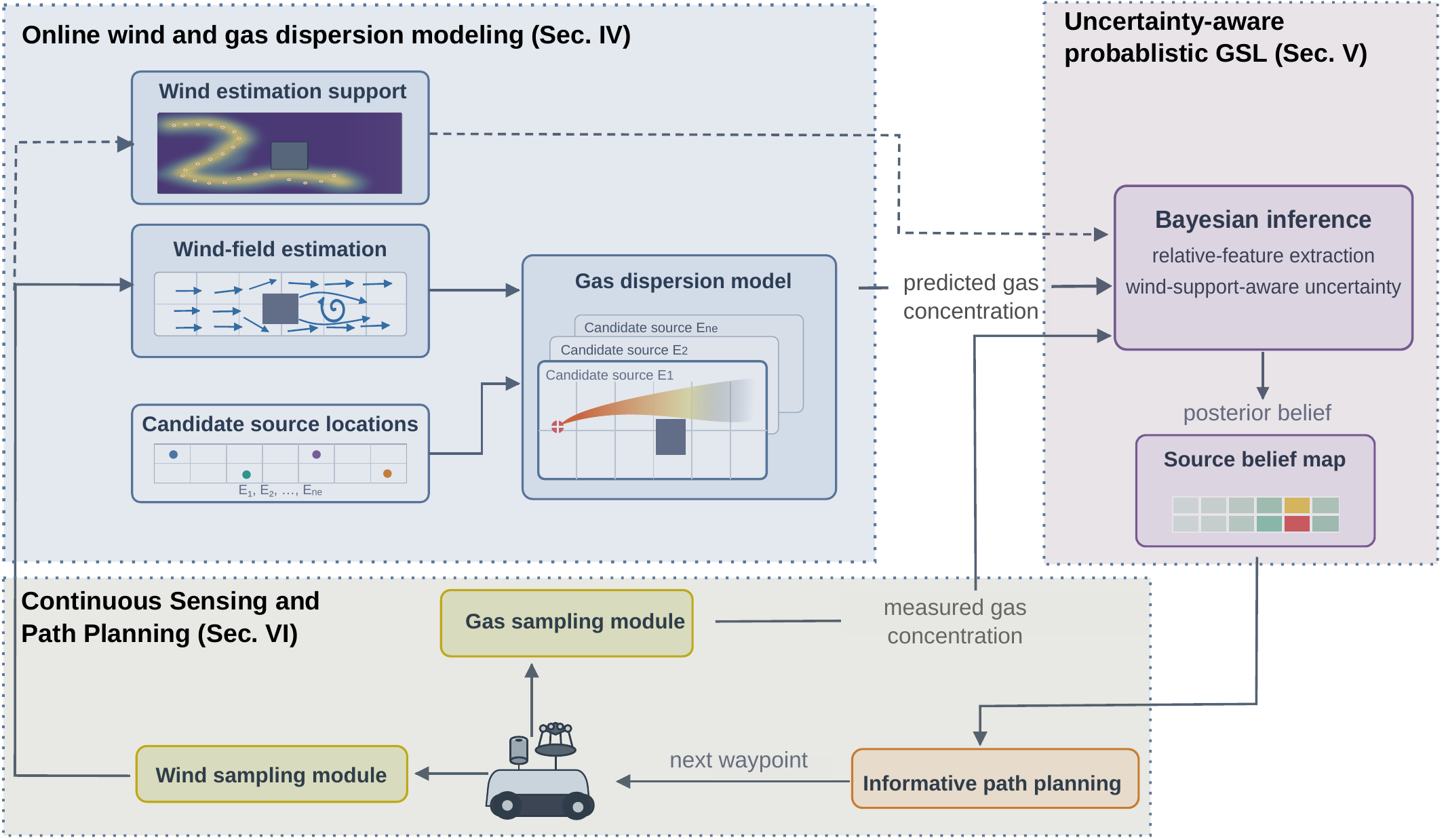}
    \caption {Pipeline of the proposed online \ac{GSL} framework. Wind and gas measurements collected during robot motion are used for online wind reconstruction, source-conditioned dispersion prediction, Bayesian belief updating, and \ac{IPP}.}
    \label{fig:STE_flowchart}
\end{figure*}

Figure~\ref{fig:STE_flowchart} provides an overview of the complete proposed
framework, including the online modeling, probabilistic inference, sensing and planning components described in Secs.~\ref{sec:wind_gas_modeling}--%
\ref{sec:continuous_sensing_planning}.

As illustrated in Fig.~\ref{fig:STE_flowchart}, the robot continuously collects wind and gas measurements while moving through the environment. After preprocessing, the wind measurements are used to reconstruct the airflow field and its local measurement support on the prediction grid. For each candidate source location, the reconstructed wind field is supplied to the \ac{ADE} solver to generate gas distribution predictions. These predictions are compared with the measured gas observations through the probabilistic inference framework. The resulting source belief guides \ac{IPP} and determines when the search terminates, thereby closing the sensing--inference--planning loop.

\section{Online Wind and Gas Dispersion Modeling} \label{sec:wind_gas_modeling}
This section describes the online wind reconstruction and forward gas dispersion prediction modules.

\subsection{Online Wind Field Reconstruction}
\label{subsec:wind_estimation}
The forward gas dispersion model requires a wind estimate over the prediction grid.  The field is reconstructed incrementally from the wind measurements using \ac{GKR}. This local, non-parametric formulation is computationally lightweight, requires minimal environmental prior, and scales well to different workspace sizes and grid resolutions.    
\subsubsection{Gaussian kernel regression}

Suppose that, up to iteration $k$, the robot has collected $N_k$ filtered wind measurements
$
    \left\{
        \left(\bar{\mathbf{z}}^w_t,\mathbf{r}_t\right)
    \right\}_{t=1}^{N_k}
$. Let $\hat{\mathbf{u}}_k:\Omega\rightarrow\mathbb{R}^2$
denote the reconstructed wind field at iteration $k$. Its value at
prediction cell $P_j$ is estimated as
\begin{equation}
    \hat{\mathbf{u}}_k
    \left(\mathbf{c}^{\mathrm{pred}}_j\right)
    =
    \frac{
        \displaystyle\sum_{t=1}^{N_k}
        w_{t,j}\bar{\mathbf{z}}^w_t
    }{
        \displaystyle\sum_{t=1}^{N_k}w_{t,j}
    },
    \label{eq:kernel_wind_estimate}
\end{equation}
where $w_{t,j}$ denotes the influence of measurement $t$ on prediction cell $j$:
\begin{equation}
    w_{t,j}
    =
    \exp\left(
        -\frac{
            D^2\left(\mathbf{r}_t,
            \mathbf{c}^{\mathrm{pred}}_j\right)
        }{2\ell_u^2}
    \right).
    \label{eq:wind_kernel_weight}
\end{equation}
Here, $\ell_u>0$ is the kernel length scale and $D(\cdot,\cdot)$ is a spatial distance measure. When no geometric information is available, we use the isotropic Euclidean distance
\begin{equation}
    D^2\left(\mathbf{r}_t,\mathbf{c}^{\mathrm{pred}}_j\right)
    =
    \left\|
        \mathbf{c}^{\mathrm{pred}}_j-\mathbf{r}_t
    \right\|_2^2 .
    \label{eq:isotropic_distance}
\end{equation}

New measurements can be incorporated incrementally by accumulating their kernel contributions to the existing wind field estimation. Consequently, the calculation complexity is linear in the number of new measurements, rather than requiring the complete wind reconstruction to be solved from scratch. 

\subsubsection{Optional obstacle-aware anisotropic kernel}
Obstacles in the environment typically distort the wind field. In particular, walls and obstacles tend to preserve flow continuity along their surface while causing stronger variation in normal direction of the wall. When obstacle geometry is available, either a priori or acquired online, it can optionally be incorporated into the kernel distance to better represent the obstacle effect. Deforming the kernel according to the local obstacle geometry increases the similarity between cells distributed along the same surface while attenuating the propagation of measurements in the surface-normal direction and across obstacle boundaries. Let $d(\mathbf{x})$ denote the signed distance to the nearest obstacle boundary and let
\begin{equation}
    \mathbf{n}_j
    =
    \frac{
        \nabla d(\mathbf{c}^{\mathrm{pred}}_j)
    }{
        \left\|
            \nabla d(\mathbf{c}^{\mathrm{pred}}_j)
        \right\|_2
    }
    \label{eq:obstacle_normal}
\end{equation}
denote the corresponding unit normal at prediction location $\mathbf{c}^{\mathrm{pred}}_j$. For the displacement
\begin{equation}
    \boldsymbol{\delta}_{t,j}
    =
    \mathbf{c}^{\mathrm{pred}}_j-\mathbf{r}_t,
\end{equation}
the wall-normal and wall-tangential components are
\begin{align}
    \boldsymbol{\delta}^{\perp}_{t,j}
    &=
    \left(
        \boldsymbol{\delta}_{t,j}^{\top}\mathbf{n}_j
    \right)\mathbf{n}_j, \\
    \boldsymbol{\delta}^{\parallel}_{t,j}
    &=
    \boldsymbol{\delta}_{t,j}
    -
    \boldsymbol{\delta}^{\perp}_{t,j}.
\end{align}

The obstacle-aware distance is defined as
\begin{equation}
    D^2\left(\mathbf{r}_t,\mathbf{c}^{\mathrm{pred}}_j\right)
    =
    \left\|
        \boldsymbol{\delta}^{\parallel}_{t,j}
    \right\|_2^2
    +
    \beta_j
    \left\|
        \boldsymbol{\delta}^{\perp}_{t,j}
    \right\|_2^2,
    \label{eq:anisotropic_distance}
\end{equation}
where
\begin{equation}
    \beta_j
    =
    1+
    \frac{\gamma}{
        |d(\mathbf{c}^{\mathrm{pred}}_j)|+\epsilon
    }.
    \label{eq:anisotropic_factor}
\end{equation}
The parameter $\gamma>0$ controls the strength of the anisotropy, and $\epsilon>0$ prevents numerical singularities near obstacle boundaries. The larger normal penalty reduces interpolation across nearby boundaries while preserving stronger correlation along them. When obstacle information is unavailable, we set $\gamma=0$, and Eq.~\eqref{eq:anisotropic_distance} reduces to the isotropic distance in Eq.~\eqref{eq:isotropic_distance}.

\subsubsection{Local measurement support}
Kernel regression is reliable only when a prediction location is sufficiently supported by nearby measurements. Let
$W_k$ denote the wind measurement support over the prediction grid, which is quantified by using the accumulated kernel weight
\begin{equation}
W_k\left(\mathbf{c}^{\mathrm{pred}}_j\right)
    =
    \sum_{t=1}^{N_k}w_{t,j}.
    \label{eq:wind_support}
\end{equation}
A large value of $W_k$ indicates strong support from nearby measurements, whereas a small value indicates that the location is undersampled. The support field $W_k$ is subsequently used as a proxy for wind estimation reliability in Sec.~\ref{subsec:wind_support_uncertainty}. 

In our setting, the wind samples are collected at irregular measurement locations $\textbf{r}_t$, whereas the wind field is reconstructed at the fixed estimation locations $\textbf{c}_j^{\mathrm{pred}}$. As a result, some prediction grid cells may lie in regions that have not yet been adequately explored by the robot. We therefore define the reconstructed wind field as
\begin{equation}
\hat{\mathbf{u}}_k
\left(\mathbf{c}^{\mathrm{pred}}_j\right)
=
\begin{cases}
\displaystyle
\frac{
    \sum_{t=1}^{N_k}w_{t,j}\bar{\mathbf{z}}^w_t
}{
    W_k(\mathbf{c}^{\mathrm{pred}}_j)
},
&
W_k(\mathbf{c}^{\mathrm{pred}}_j)\geq\tau_w, \\[3mm]
\mathbf{u}^{\mathrm{bg}}_k,
&
W_k(\mathbf{c}^{\mathrm{pred}}_j)<\tau_w,
\end{cases}
\label{eq:supported_wind_estimate}
\end{equation}
When the wind support at a prediction cell falls below the threshold $\tau_w>0$, the locally reconstructed wind estimate is replaced by the background wind estimate $\mathbf{u}^{\mathrm{bg}}_k$. This avoids numerical instability in unexplored regions.

\subsubsection{Background wind estimation}

The background wind is used to approximate the dominant wind field in regions not yet supported by local measurements, to provide a rough heuristic for source estimation. Define the mean of the wind measurement gathered till iteration $k$ as
\begin{equation}
    \bar{\mathbf{u}}^w_k
    =
    \frac{1}{|N_k|}
    \sum_{t\in N_k}
    \bar{\mathbf{z}}^w_t.
    \label{eq:batch_wind_mean}
\end{equation}
The background estimate is updated using an exponential moving
average:
\begin{equation}
    \mathbf{u}^{\mathrm{bg}}_k
    =
    (1-\alpha)\mathbf{u}^{\mathrm{bg}}_{k-1}
    +
    \alpha\bar{\mathbf{u}}^w_k,
    \label{eq:background_wind}
\end{equation}
where $\alpha\in(0,1]$ controls the update rate. As exploration progresses, local kernel estimates become available over a larger portion of the prediction grid, and the influence of the background estimate is correspondingly reduced.

\subsection{ADE-based Gas Dispersion Prediction}\label{subsec:gas_plume_model}
The reconstructed wind field is used to predict the time-averaged gas distribution pattern associated with each candidate source location. 

\subsubsection{Governing model}

For each source hypothesis $E_i$, let
$C_k(\mathbf{x}\mid E_i)$ denote the predicted gas concentration and $\hat{\mathbf{u}}_k(\mathbf{x})$ denote the reconstructed wind field
at position $\mathbf{x}\in\Omega$ and iteration $k$. For clarity, we consider a fixed inference iteration $k$ and source hypothesis
$E_i$ throughout the following derivation. We write
$
C(\mathbf{x}) \equiv \hat{C}_k(\mathbf{x}\mid E_i),
$ and $
\mathbf{u}(\mathbf{x}) \equiv \hat{\mathbf{u}}_k(\mathbf{x}),
$ and omit the iteration
and source-hypothesis indices. Assuming a spatially uniform effective diffusion coefficient $\kappa$, the concentration field satisfies
\begin{equation}
    -\kappa\nabla^2
    C_(\mathbf{x})
    +
    \hat{\mathbf{u}}(\mathbf{x})^\top
    \nabla C(\mathbf{x})
    =
    q_i(\mathbf{x}),
    \label{eq:continuous_ade}
\end{equation}

where $q_i(\mathbf{x})$ is the source term associated with $E_i$. The effective diffusion coefficient $\kappa$ represents unresolved turbulent mixing and controls the lateral spread of the predicted plume. The boundary conditions are detailed in Sec.~\ref{sec:boundary_condition}.

\subsubsection{Finite difference discretization}
Eq.~\eqref{eq:continuous_ade} is solved numerically on the fine prediction grid using the \ac{FDM}. \ac{FDM} approximates spatial derivatives of \ac{PDE} using differences between neighboring grid points, which converts the differential equations into a system of algebraic equations. 
Let the $\Delta x$ and $\Delta y$ define the prediction grid spacings, and $C_{m,n}$ and $\hat{\mathbf{u}}_{m,n} =[\hat{u}_{x,m,n},\hat{u}_{y,m,n}]^\top$
denote the concentration and reconstructed wind vector at prediction cell $(m,n)$, respectively. Diffusion is discretized using second-order central differences.
\begin{align}
    \frac{\partial^2 C}{\partial x^2} &\approx \frac{C_{m+1,n} - 2C_{m,n} + C_{m-1,n}}{\Delta x^2}, \\
    \frac{\partial^2 C}{\partial y^2} &\approx \frac{C_{m,n+1} - 2C_{m,n} + C_{m,n-1}}{\Delta y^2}.
\end{align}
To improve numerical stability, the advection is approximated using a first-order upwind scheme.
\begin{align}
    \frac{\partial C}{\partial x} &\approx 
    \begin{cases}
        \dfrac{C_{m,n} - C_{m-1,n}}{\Delta x}, & \hat{u}_{x,m,n} > 0, \\[1em]
        \dfrac{C_{m+1,n} - C_{m,n}}{\Delta x}, & \hat{u}_{x,m,n} < 0,
    \end{cases} \\[1em]
    \frac{\partial C}{\partial y} &\approx 
    \begin{cases}
        \dfrac{C_{m,n} - C_{m,n-1}}{\Delta y}, & \hat{u}_{y,m,n} > 0, \\[1em]
        \dfrac{C_{m,n+1} - C_{m,n}}{\Delta y}, & \hat{u}_{y,m,n} < 0.
    \end{cases}
\end{align}
The discrete equation at an interior cell is written as
\begin{equation}
\begin{split}
    a_P C_{m,n}
    -a_W C_{m-1,n}
    -a_E C_{m+1,n}
    \\-a_S C_{m,n-1}
    -a_N C_{m,n+1}
    =
    q_{m,n}^{(i)},
\end{split}
\label{eq:ade_stencil}
\end{equation}
where
\begin{align}
    a_W &=
    \frac{\kappa}{\Delta x^2}
    +
    \frac{\max(\hat{u}_{x,m,n},0)}{\Delta x},\\
    a_E &=
    \frac{\kappa}{\Delta x^2}
    +
    \frac{\max(-\hat{u}_{x,m,n},0)}{\Delta x},\\
    a_S &=
    \frac{\kappa}{\Delta y^2}
    +
    \frac{\max(\hat{u}_{y,m,n},0)}{\Delta y},\\
    a_N &=
    \frac{\kappa}{\Delta y^2}
    +
    \frac{\max(-\hat{u}_{y,m,n},0)}{\Delta y},\\
    a_P &= a_W+a_E+a_S+a_N.
\end{align}

For source hypothesis $E_i$, let $(m_i,n_i)$ denote the nearest prediction grid cell to its center $\mathbf{c}^{\mathrm{est}}_i$. The discrete source term is
\begin{equation}
    q_{m,n}^{(i)}
    =
    \begin{cases}
        \displaystyle\frac{Q}{\Delta x\Delta y},
        & (m,n)=(m_i,n_i),\\[2mm]
        0, & \text{otherwise},
    \end{cases}
    \label{eq:discrete_source}
\end{equation}
where $Q$ denotes the total source strength. Scaling the source by the cell area preserves the total release strength across different prediction grid resolutions.

\subsubsection{Boundary Condition} \label{sec:boundary_condition}
In our work, we assume that the candidate source region is sufficiently far from the outer boundary of the computational domain for boundary effects on the predicted plume pattern to remain limited. a homogeneous
Dirichlet condition is applied to the source-conditioned \ac{ADE} prediction
along the outer boundary:
\begin{equation}
    C(\mathbf{x})=0,
    \qquad \mathbf{x}\in\partial\Omega
    \label{eq:default_outer_boundary}
\end{equation}
Here, $\partial\Omega$ denotes the outer boundary of the computational domain. This condition treats the outer boundary as a numerical truncation of the modeled environment. 

No explicit boundary condition is imposed on internal obstacles or walls in the default formulation. Instead, the reconstructed wind field implicitly conveys their dominant influence on advective gas transport. Wind measurements collected around an obstacle reflect local changes in flow direction and magnitude, which are subsequently incorporated into the \ac{ADE} model. This is a task-oriented approximation: it captures obstacle-induced advection but does not strictly prevent numerical diffusion through an unmodeled solid boundary. Nevertheless, the subsequent inference compares the relative spatial patterns of the predicted and measured gas signals, rather than their absolute concentration values, as detailed in
Sec.~\ref{subsec:relative_gas_pattern}. This reduces the sensitivity of the localization process to local concentration errors introduced by the simplified boundary treatment. Therefore, the proposed framework can operate without prior knowledge of the environmental layout, obstacle geometry, or inlet and outlet locations.

When the environmental layout is known a priori or acquired online during robot exploration, explicit physical boundary conditions can instead be incorporated to improve the physical fidelity of the predicted gas field. A homogeneous Dirichlet condition can be applied at inlet cells, assuming that no gas enters the domain from outside. Impermeable walls and obstacle surfaces can be assigned a zero-normal-flux Neumann condition, while a zero-normal-gradient condition can be applied at outlet cells. Thus, the model remains operational with limited prior environmental information while allowing additional geometric and boundary information to be incorporated when available.

\subsubsection{Sparse matrix assembly} \label{subsec:ade_matrix_assembly}
After discretizing the \ac{ADE} over the prediction grid and applying the boundary conditions, the equations are assembled into the sparse linear system
\begin{equation}
    \mathbf{A}_k\mathbf{C}^{(i)}_k
    =
    \mathbf{q}^{(i)},
    \label{eq:ade_linear_system}
\end{equation}
where $\mathbf{C}^{(i)}_k\in\mathbb{R}^{N_p}$ contains the predicted gas concentration values over the prediction grid for source hypothesis $E_i$.
The $\mathbf{q}^{(i)}\in\mathbb{R}^{N_p}$ is the corresponding discrete source vector, and $\mathbf{A}_k\in\mathbb{R}^{N_p\times N_p}$ is the sparse coefficient matrix determined by the reconstructed wind field, effective diffusion coefficient, grid resolution, and boundary treatment.

The linear system is solved using the SuperLU sparse direct solver through the Eigen interface. Within iteration $k$, all source hypotheses share the same coefficient matrix $\mathbf{A}_k$ since the reconstructed wind field and model parameters remain fixed. As a result, the computationally expensive factorization of $\mathbf{A}_k$ is performed only once per iteration. The resulting factorization is reused to solve Eq.~\eqref{eq:ade_linear_system} for each right-hand side $\mathbf{q}^{(i)}$ that depends on the source. When the wind field is updated at the next iteration, $\mathbf{A}_k$ is reassembled and factorized again.

\subsubsection{Model parameter selection}
\label{subsec:ade_parameter_selection} The \ac{ADE} model contains two main scalar parameters: the effective diffusion coefficient $\kappa$ and the source strength $Q$. The effective diffusion coefficient represents unresolved turbulent mixing and controls the spatial spread of the predicted plume. A small value of $\kappa$ produces a narrow, advection-dominated plume, whereas a larger value produces greater lateral spreading.

\begin{figure}
    \centering
    \includegraphics[width=0.8\linewidth]{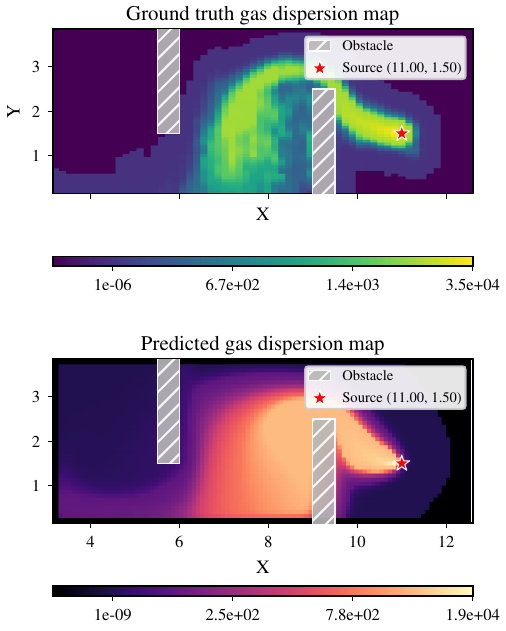}
    \caption{Representative comparison between the gas distribution map
generated by the filament-based simulation and the steady-state \ac{ADE}
prediction for the same source location and wind field.}
    \label{fig:filament_vs_ADE}
\end{figure}

We select $\kappa$ by comparing the \ac{ADE} predictions with reference plumes generated by the filament-based gas model. The value providing the best agreement in plume shape is selected for the subsequent localization experiments. Illustrated by the representative comparison in Fig.~\ref{fig:filament_vs_ADE}, $\kappa=0.008$ captures the dominant transport direction and overall plume structure sufficiently well for relative pattern that is used for source inference.

The second model parameter is the source strength $Q$, which is generally unknown in practical applications. Previous model-based approaches either assume the release rate as prior information~\cite{jin_towards_2023} or include it as an additional source parameter to be estimated~\cite{rahbar2019algorithm}. The latter increases the dimensionality of the inference problem. In this work, we instead compare relative gas patterns, as detailed in Sec.~\ref{subsec:relative_gas_pattern}. Because the source strength $Q$ scales the magnitude of the predicted concentration field without changing its normalized spatial pattern, its exact value does not need to match the physical release rate. We therefore use the same fixed positive value of $Q$ for all source hypotheses. This design also avoids direct calibration among the physical release rate, gas sensor output, and forward dispersion model.

Table~\ref{tab:ade_computation_time} reports the total computation time required to generate all source-conditioned gas dispersion predictions for all $N_e$ candidate source locations on an Intel Core i7-12700H computer. The $100\times40$ prediction grid used in our experiments provides a suitable compromise between spatial resolution and online computational efficiency.

\begin{table}[t]
    \centering
    \caption{Total computation time required by the \ac{ADE}-based forward model to generate gas-dispersion predictions for all $N_e$ (400 in our setup) candidate source locations at different prediction-grid resolutions.}
    \label{tab:ade_computation_time}
    \begin{tabular}{cc}
        \hline
        Grid resolution & Computation time (ms) \\
        \hline
        $50 \times 20$  & 50  \\
        $100 \times 40$ & 160 \\
        $200 \times 80$ & 810 \\
        \hline
    \end{tabular}
\end{table}

\section{Uncertainty-Aware Probabilistic GSL}\label{sec:probabilistic_gsl}

This section presents the probabilistic framework used to infer the source location from the measured gas concentration and the source-conditioned dispersion predictions. It introduces the relative gas feature representation, Bayesian likelihood, and wind-support-aware uncertainty.

\subsection{Relative Gas Feature Representation} \label{subsec:relative_gas_pattern}
Model-based \ac{GSL} commonly evaluates a candidate source by comparing measured gas concentrations with the values predicted by dispersion models \cite{rahbar_design_2019,hutchinson2018information,zhu_novel_2020, jakkala_probabilistic_2022}. However, the raw output of a physical gas sensor may differ from the modeled concentration because of sensor calibration, source strength, environmental conditions, and sensor dynamics. Direct value comparison would therefore require an accurate mapping between the sensor output and the prediction of the \ac{ADE} model.

In~\cite{jin_uncalibrated_2026}, the author demonstrates that the spatial variation of the gas signal contains enough information for \ac{GSL} tasks. Inspired by this work, we compare the relative spatial variation of the measured and predicted gas signals through z-score normalization. However, it is difficult to get access to the gas concentration distribution before the robot exhaustively explores the whole environment. Therefore, we normalize the measured and predicted gas concentration with a dynamic window.

At iteration $k$, suppose that the robot has collected $N_k$ gas measurements
$\{(z^g_t,\mathbf{r}_t)\}_{t=1}^{N_k}$. For a given source hypothesis $E_i$, the \ac{ADE} model predicts the corresponding gas concentration field $\hat{C}_k(\cdot|E_i)$. To extract the model prediction corresponding to each measurement location, each sampling position $\mathbf{r}_t$ is associated with its nearest prediction-grid cell $j_t$:
\begin{equation}
    j_t
    =
    \underset{j\in\{1,\ldots,N_p\}}{\arg\min}
    \left\|
        \mathbf{r}_t-\mathbf{c}^{\mathrm{pred}}_j
    \right\|_2 .
    \label{eq:measurement_projection}
\end{equation}
The corresponding predicted gas concentration series is therefore $\{\hat{C}_k(j_t|E_i)\}_{t=1}^{N_k}$. For both the predicted and measured gas value sequences, we compute the mean values $\mu^g_k$ and $\mu^C_{k,i}$ and the standard deviation $\sigma^g_k$ and $\sigma^C_{k,i}$, respectively. We compute the normalized gas value measurements $\tilde{z}^g_t$ using z-score normalization.
\begin{equation}
    \tilde{z}^g_{k,t}
    =
    \frac{
        z^g_t-\mu^g_k
    }{
        \max(\sigma^g_k,\epsilon)
    },
    \label{eq:normalized_gas_measurement}
\end{equation}
where $\epsilon>0$ prevents numerical instability when the measured signal has negligible variation. The returned $\tilde{z}^g_{k,t}$ represents the relative distance of each measurement to the mean value of all the available measurements that have been gathered so far. Similarly, for source hypothesis $E_i$, the normalized prediction is
\begin{equation}
    \tilde{C}_{k,t
    }(E_i)
    =
    \frac{
        \hat{C}_{k}(j_t|E_i)-\mu^C_{k,i}
    }{
        \max(\sigma^C_{k,i},\epsilon)
    }.
    \label{eq:normalized_gas_prediction}
\end{equation}

This normalization removes constant offsets and positive scaling, allowing the inference to focus on the similarity between the relative spatial patterns of the measured and predicted gas values. Consequently, the physical gas sensor does not need to be calibrated directly against the concentration values produced by the \ac{ADE} model.

Because new measurements change the statistics of the complete sequences, the normalized measurements and predictions are recomputed at every inference iteration. The predicted sequences are also updated using the latest reconstructed wind field.

\subsection{Bayesian Source inference} \label{subsec:bayesian_inference}
At each iteration, the source belief is evaluated using all measurements $\mathcal{Y}_{1:k}$ collected up to that point. Since the measured and predicted gas sequences are renormalized whenever new measurements become available, the posterior is recomputed from the initial prior rather than updated recursively from $b_{k-1}$. This avoids repeatedly incorporating measurements whose normalized values have changed.

As defined in Eq.~\eqref{eq:belief_map}, $b_k(i)$ denotes the posterior belief associated with the source hypothesis $S=i$. As introduced in Sec.~\ref{sec:problem_statement}, the prior $b_0(i)$ is initialized as a uniform distribution over the free space of the environment, while cells occupied by obstacles are assigned zero probability. The posterior is evaluated as
\begin{equation}
    b_k(i) \propto \mathcal{L}_k(i)b_0(i),
    \qquad i=1,\ldots,N_e, 
\end{equation}
followed by normalization such that $\sum_{i=1}^{N_e} b_k(i)=1.$ where  $\mathcal{L}_k(i)$ denotes the likelihood score of source hypothesis $E_i$.

Using the normalized gas features introduced in
Sec.~\ref{subsec:relative_gas_pattern}, the discrepancy between the measured and predicted relative gas
patterns for source hypothesis $E_i$ is defined as
\begin{equation}
    \mathrm{Diff}_k(E_i)
    =
    \sum_{t=1}^{N_k}
    \left(
        \tilde{C}_{k,t}(E_i)
        -
        \tilde{z}^{g}_{k,t}
    \right)^2.
    \label{eq:pattern_discrepancy}
\end{equation}

We model the normalized measurement and predicted gas features, $\{\tilde{z}^{g}_{k,t}\}_{t=1}^{N_k}$ and ${\{\tilde{C}_{k,t}(E_i)}\}_{t=1}^{N_k}$, as noisy observations of an underlying relative gas pattern. Their errors are approximated by independent zero-mean Gaussian distributions. The measurement error variance $\sigma_g^2$ represents slow sensor dynamics, sensor noise, motion-induced disturbances, and other unmodeled effects in the sensing process. The prediction error variance $\sigma_{C,k}^2$ represents uncertainty in the gas dispersion prediction caused by model mismatch and unmodeled physical effects.

Based on these Gaussian error assumptions and assuming that the measurement and prediction errors are independent, we define the generalized likelihood score as:
\begin{equation}
    \mathcal{L}_k(i)
    =
    \exp\left(
        -\frac{
            \mathrm{Diff}_k(E_i)
        }{
           2( \sigma_g^2+\sigma_{C,k}^2(E_i))
        }
    \right),
    \label{eq:tempered_pattern_likelihood}
\end{equation}

The sum $\sigma_g^2+\sigma_{C,k}^2(E_i)$ controls the sensitivity of the likelihood score to differences between the measured and predicted relative gas patterns. A smaller value suppresses inconsistent source hypotheses more strongly, whereas a larger value produces a flatter likelihood score.

In conventional formulations, $\sigma_g^2$ and $\sigma_{C,k}(E_i)^2$ are typically selected empirically and applied uniformly to all source hypotheses through all the iterations \cite{ojeda_robotic_2023,jin_cumulative_2025}. In our setting, however, the wind field is reconstructed incrementally from measurements collected during exploration. Consequently, the reliability of the source-conditioned gas predictions varies both spatially and temporally. We therefore model the additional prediction uncertainty using the local wind-support evidence introduced in Sec.~\ref{subsec:wind_support_uncertainty}.

\subsection{Wind-Support-Aware Likelihood}
\label{subsec:wind_support_uncertainty}
As the measurements are collected along the robot trajectory, regions close to the collected measurements are generally better supported than unexplored regions \cite{Thomas2010}\cite{monroy_online_2017}. Treating all source-conditioned predictions as equally reliable can therefore produce an overconfident belief during the early stages of exploration.

A poorly supported wind estimate at the candidate source affects the predicted plume from its point of release and can therefore produce an unreliable gas-distribution pattern. We use the local wind support at the candidate source as a computationally lightweight proxy for this effect. Let $W_k(E_i)$ denote the accumulated wind measurement support defined in Eq.~\eqref{eq:wind_support}, evaluated at the prediction-grid cell nearest to the center $\mathbf{c}^{\mathrm{est}}_i$ of source hypothesis $E_i$. The corresponding wind-support-aware uncertainty is defined as

\begin{equation}
    \sigma_{C,k}^2(E_i)
    = \sigma_0^2 + \lambda_k^w(E_i)
    \label{eq:wind_support_uncertainty}
\end{equation}
with
\begin{equation}
    \lambda_k^w =A_w\exp\left[-W_k(E_i)\right].    
\end{equation}
    
Here, $\sigma_0^2 >0$ denotes the baseline prediction error variance associated with mismatch between the \ac{ADE} approximation and the physical gas distribution when the reconstructed wind field is sufficiently supported. The term $\lambda_k^w(E_i)$ represents the additional prediction uncertainty attributed to limited wind measurement support near the candidate source $E_i$. The parameter $A_w\geq0$ specifies the maximum additional prediction-error variance in the absence of local wind support. Accordingly, $\lambda_k^w(E_i)\in(0, A_w]$ and decreases monotonically as the local measurement support increases.

When the wind estimate near candidate source $E_i$ is poorly supported, $W_k(E_i)$ is small and the corresponding $\sigma_C^2(E_i)$ increases. Through Eq.~\eqref{eq:tempered_pattern_likelihood}, this increases the tolerance to discrepancies between the measured and predicted relative gas patterns and therefore prevents the corresponding source hypothesis from being overly penalized. As additional wind measurements are collected near the candidate source, $\lambda_k^w(E_i)$ decreases toward zero and the likelihood gradually recovers the baseline sensitivity determined by $\sigma_0^2$ and $\sigma_g^2$.

\section{Continuous Sensing and Path Planning}\label{sec:continuous_sensing_planning}
This section describes how the robot gathers measurements and uses the inferred source belief to close the sensing--planning loop.

\subsection{Continuous Sensing and Preprocessing} \label{sec:sensing_preprocessing}
Unlike stop–sense–go strategies commonly used in existing works \cite{ojeda_robotic_2023}\cite{wiedemann_physics-informed_2025}, in which the robot stops at selected locations before recording measurements, the proposed system collects wind and gas measurements continuously during motion. Although individual measurements may be affected by robot motion and sensor dynamics, continuous sensing substantially increases the spatial sampling density and reduces the time spent waiting at discrete sensing locations \cite{jin_sense_2024}. Each measurement is time-stamped and associated with the robot position at which it was acquired.

\subsubsection{Wind measurement}
Let $\mathbf{z}^{a}_{t}\in\mathbb{R}^{2}$ denote the planar velocity measured by the anemometer at sample $t$, expressed in the anemometer frame. The measurement is first transformed into the world frame. Let $\mathbf{R}^{r}_{a}$ denote the rotation from the anemometer frame to the robot frame and let $\mathbf{R}^{w}_{r,t}$ denote the rotation from the robot frame to the world frame at sample $t$. The measured relative wind velocity in the world frame is
\begin{equation}
    \mathbf{z}^{w,\mathrm{rel}}_{t}
    =
    \mathbf{R}^{w}_{r,t}
    \mathbf{R}^{r}_{a}
    \mathbf{z}^{a}_{t}.
    \label{eq:wind_frame_transform}
\end{equation}
Since the anemometer is mounted on a moving robot, it measures the airflow velocity relative to the robot rather than the environmental wind velocity. The robot's translational motion introduces an airflow in the direction opposite to its motion. To compensate for this effect, we denote the robot translational velocity in the world frame as $\mathbf{v}^{w}_{r,t}\in\mathbb{R}^{2}$. The environmental wind velocity can then be estimated as
\begin{equation}
    \mathbf{z}^{w}_{t}
    =
    \mathbf{z}^{w,\mathrm{rel}}_{t}
    +
    \mathbf{v}^{w}_{r,t}.
    \label{eq:wind_motion_compensation}
\end{equation}

Finally, a moving average filter is applied to reduce high frequency measurement noise using the most recent $N_w$ samples. Filtered samples
$\{(\bar{\mathbf{z}}^{w}_{t},\mathbf{r}_{t})\}$ are used for the online wind field reconstruction.

\subsubsection{Gas measurement}
The gas sensor output $z^{g}_{t}$ is recorded continuously and associated with
the corresponding robot position $\mathbf{r}_{t}$. As described in
Sec.~\ref{subsec:relative_gas_pattern}, the proposed inference method compares
the relative spatial patterns of the measured and predicted gas signals rather
than their absolute values. The raw sensor output is therefore used directly
for relative pattern extraction without further processing. This eliminates the need for sensor calibration and therefore facilitates efficient deployment with different gas sensors and source strengths.

\subsection{Informative Path Planning} \label{sec:IPP}
At each iteration, the current posterior belief is used to select the next region for exploration. The objective of the \ac{IPP} strategy is to select robot trajectories that collect informative measurements for subsequent belief updates, thereby accelerating the localization process. Because wind and gas measurements are acquired continuously during robot motion, the planner should favor trajectories that traverse informative regions rather than only selecting isolated informative sampling locations.

We adapt the belief-clustering strategy introduced in~\cite{jin_sense_2024}, using the posterior probability $b_k(i)$ as the informativeness score of estimation cell $E_i$ ~\cite{jin_cumulative_2025}. This greedy belief feature prioritizes regions with a high probability of containing the source. The planner first identifies informative cells according to their posterior probability and then groups spatially adjacent informative cells into regions. After all the informative regions are defined, we select the closest one and define a path to scan through it. 

To characterize how concentrated the current source belief is, we use its Shannon entropy:

\begin{equation}
    H(b_k)
    =
    -\sum_{i=1}^{N_e}
    b_k(i)\log b_k(i),
    \label{eq:posterior_entropy}
\end{equation}
A high entropy indicates that the posterior mass is distributed across multiple source hypotheses, whereas a low entropy indicates that the belief is concentrated in a smaller region. At each iteration, informative cells are selected from the current belief map using an adaptive threshold based on the posterior entropy:

\begin{equation}
 \rho_k = 1 - \frac{H(b_k)}{\zeta} \quad \text{with} \quad \zeta\geq H(b_0),
\end{equation}
where $b_0$ is the initial uniform belief. Once the robot starts to explore the environment, the resulting $H(b_k)$ will always be smaller than $H(b_0)$, which results in $\rho_k\in(0,1]$. The corresponding threshold is defined as:
\begin{equation}
    \theta_k = \rho_k * (\max_i b_k(i) - \min_i b_k(i)) + \min _ib_k(i)
\end{equation}

All estimation grid cells satisfying $b_k(i)>\theta_k$ are treated as informative cells. This adaptive threshold controls a trade-off between the exploration-exploitation behavior of the planner. Early in the search, when only limited measurements are available and the posterior remains diffuse, the threshold is low and a broader set of candidate regions is retained. As additional information is collected and the posterior becomes concentrated, the threshold increases, focusing exploitation on the most probable source regions.

The selected informative cells are then grouped into spatial clusters using DBSCAN~\cite{ester1996density}. Rectangle bounding boxes are defined for each cluster by identifying the spread range of each cluster. The cluster closest to the robot is selected to be visited first, and the bounding-box vertex farthest from the current robot position is chosen as the next goal. This encourages the robot trajectory to traverse the selected informative region. If no valid cluster remains, the cell with the highest informativeness score is selected as the next goal.

\subsection{Termination Condition}
\label{subsec:termination}

The robot continuously selects informative goals until either the source belief is sufficiently concentrated or the maximum number of iterations is reached. The source is declared when $H(b_k)$ falls below the prescribed threshold $H_{\mathrm{th}}$. If this condition is not satisfied, exploration continues until the maximum number of iterations $K_{\max}$ is reached. In all cases, the declared source location is given by the maximum a posteriori estimate defined in Eq.~\eqref{eq:map_source_estimate}. The values of $H_{\mathrm{th}}$ and $K_{\max}$ are reported in the corresponding experimental setup.

\section{Simulation Experiments}  \label{sec:simulation_exp}
The simulation experiments evaluate the proposed framework progressively while separating the effect of path planning from that of source inference. We first evaluate the source estimation methods using a predefined coverage trajectory, ensuring that all methods receive measurements along the same path. Under this setting, we first compare the gas dispersion models using the ground truth wind field and subsequently evaluate the effect of reconstructing the wind field online. Finally, each source estimation method is integrated with the \ac{IPP} strategy to evaluate the complete closed-loop \ac{GSL} system.

\subsection{Simulation Setup} \label{subsec: simulation_setup}
In this part, we introduce the simulation setup with the simulation tools, testing maps and evaluation metrics.

\subsubsection{Simulation tools}
\begin{figure}[t]
\vspace{0.2cm}
    \centering
    \includegraphics[width=0.9\linewidth]{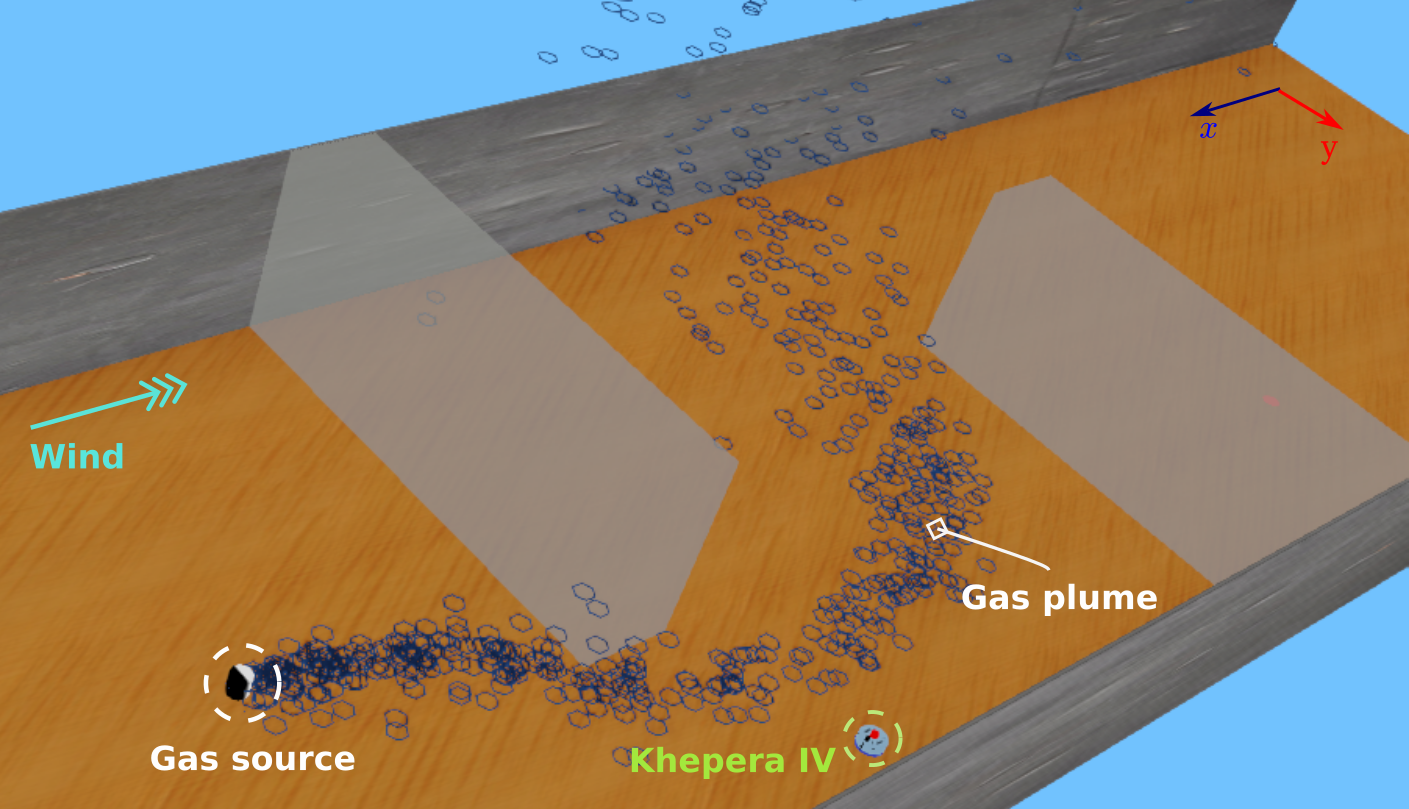}
    \caption{High-fidelity simulation environment implemented in Webots, showing the mobile robot, gas source, obstacle geometry, and gas filaments transported by the OpenFOAM-generated airflow field.}
    \label{fig: Webots_setup}
\end{figure}
The experiments are conducted in Webots~\cite{Webots} using a filament-based gas dispersion plugin~\cite{webots_odor}. For each environment, a stationary reference wind field is generated in OpenFOAM using the prescribed geometry and boundary conditions, imported into Webots to transport the gas filaments, and used as ground truth for evaluation. Although the wind field is stationary,
the simulated gas dispersion is dynamic and intermittent, causing the concentration at each location to vary over time. Each combination of environment and localization method is evaluated over 30 independent runs. Fig.~\ref{fig: Webots_setup} illustrates one of the simulated scenarios in Webots. To evaluate robustness across source configurations, the source position is randomized between runs. For the evaluation of the gas dispersion and online wind reconstruction modules, the robot follows the same predefined lawnmower trajectory in every run, starting near one end of the $x$-axis and traversing the environment with a spacing of $0.5\,\mathrm{m}$ between adjacent scan lines, which results in a total length around 80 m. For experiments integrating source estimation with \ac{IPP}, the robot's initial position is also randomized, and each run terminates when one of the stopping conditions defined in Sec.~\ref{subsec:termination} is satisfied.

\subsubsection{Sensor simulation}
The simulated gas sensor approximates the \ac{PID} used in the physical experiments. 
Sensor response and recovery dynamics are represented using two first-order systems, with a response time $\tau_{\mathrm{rise}}$ and recovery time $\tau_{\mathrm{recovery}}$. Since the proposed inference uses relative gas patterns, the localization algorithm receives the raw simulated sensor output without access to the calibration parameters. The wind sensor is simulated by sampling the ground truth wind vector at the robot position. The resulting measurements are processed using the motion compensation and filtering procedure described in Sec.~\ref{sec:sensing_preprocessing}.

\subsubsection{Testing scenarios}

\begin{figure}
    \centering
    \includegraphics[width=\linewidth]{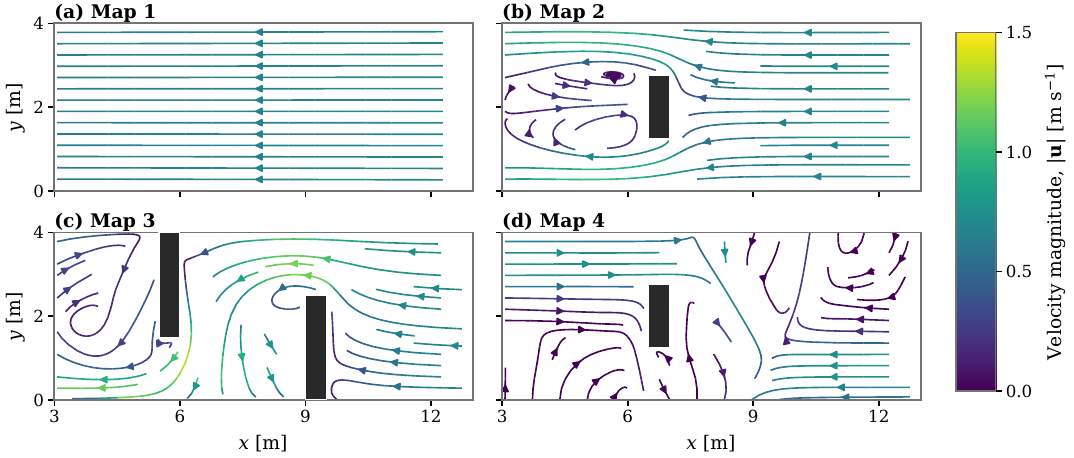}
    \caption{Ground-truth wind fields of the four simulated environments: (a) obstacle-free flow, (b) flow around a single obstacle, (c) flow around multiple obstacles, and (d) multidirectional airflow. Arrows indicate the local wind direction, and colors represent the wind-speed magnitude.}
    \label{fig:wind_field_simulated_maps}
\end{figure}

We consider four $10\,\mathrm{m}\times4\,\mathrm{m}$ environments with increasing spatial airflow complexity, as shown in Fig.~\ref{fig:wind_field_simulated_maps}.

\begin{itemize}
    \item \textbf{Map 1 (Obstacle-free flow):} An obstacle-free environment with a dominant flow from right to left. This map approximately satisfies the spatially uniform wind assumption used by analytical plume models.

    \item \textbf{Map 2 (Single obstacle):} A $0.5\,\mathrm{m}\times2\,\mathrm{m}$ obstacle is placed within the dominant flow, producing flow separation and a disturbed wake.

    \item \textbf{Map 3 (Multiple obstacles):} Two large obstacles partially block the dominant flow and produce interacting wake regions and stronger spatial variations in wind direction and magnitude.

    \item \textbf{Map 4 (Multidirectional flow):} Multiple airflow inlets generate different flow directions around
    an internal obstacle, making the wind direction change rapidly across regions.  
\end{itemize}

\subsubsection{Evaluation Metrics}
Localization performance is evaluated using the following metrics:

\begin{itemize}
    \item \textbf{Localization error:}
    the Euclidean distance between the true source position $\mathbf{x}_s$ and the declared source position $\hat{\mathbf{x}}_s$.

    \item \textbf{Success rate:}
    a run is considered successful when $|\hat{x}_s-x_s|\leq1\,\mathrm{m}$ and $|\hat{y}_s-y_s|\leq0.4\,\mathrm{m}$, which is defined by two-cell distance of the estimation grid.

    \item \textbf{Travel distance:}
    The distance traveled from the initial robot position until termination.

\end{itemize}

\newcommand{\placeholder}{\mbox{}}

\begin{table*}[t]
  \centering
  \caption{Summary of the parameters used in the simulation experiments.}
  \label{tab:simulation_parameters}
  \renewcommand{\arraystretch}{1.12}
  \setlength{\tabcolsep}{5pt}
  \begin{tabularx}{\textwidth}{@{}>{\raggedright\arraybackslash}p{0.18\textwidth}
                                      >{\raggedright\arraybackslash}X
                                      >{\centering\arraybackslash}p{0.18\textwidth}@{}}
    \toprule
    Label & Meaning & Value \\
    \midrule
    \multicolumn{3}{@{}l}{\textit{Environment and discretization}} \\
    $L_x \times L_y$ & Dimensions of the simulated search environment (m) & 10  $\times$ 4  \\
    $N_e$ & Number of cells in the source-estimation grid & 400 \\
    $N_p$ & Number of cells in the wind/gas prediction grid & 4000 \\
    $\Delta x,\,\Delta y$ & Prediction-grid spacing along the two spatial axes (m) & 0.1  $\times$ 0.1 \\
    \midrule
    \multicolumn{3}{@{}l}{\textit{Wind-field reconstruction}} \\
    $\ell_u$ & Length scale of the Gaussian kernel used for wind reconstruction (m)&  0.3 \\
    $\tau_w$ & Minimum accumulated kernel support required to use the local wind estimate & $10^{-6}$\\
    $\alpha$ & Update coefficient of the background-wind estimate & 0.9 \\
    $\gamma$ & Strength of the optional obstacle-aware kernel anisotropy & 0.5 \\
    $\epsilon$ & Small positive constant used for numerical stability & $10^{-6}$ \\
    $N_w$ & Window length (or number of samples) used by the wind-measurement filter & 10 \\
    \midrule
    \multicolumn{3}{@{}l}{\textit{Gas-dispersion prediction and inference}} \\
    $\kappa$ & Effective diffusion coefficient in the ADE forward model & 0.008 \\
    $Q$ & Nominal source strength used for all source hypotheses & 1500 \\
    $\sigma_g^2 + \sigma_0^2$ & Measurement-error variance and baseline gas-prediction error variance & 2.5 \\
    $A_w$ & Maximum additional prediction-error variance due to limited wind support & 1 \\
    $\tau_{\mathrm{rise}},\,\tau_{\mathrm{recovery}}$ & Time constants of the simulated gas-sensor response and recovery dynamics (s) & 2.04 s, 4.57 s \\
    \midrule
    \multicolumn{3}{@{}l}{\textit{Robot motion, planning, and stopping}} \\
    $v_{\max}$ & Maximum robot translational speed (m/s) & 0.27 \\
    $f_s$ & Sampling frequency of the simulated gas and wind sensors (Hz) & 10 \\
    $H_{\mathrm{th}}$ & Posterior-entropy threshold for declaring the source & 0.1 \\
    $K_{\max}$ & Maximum number of source-estimation iterations & 30 \\
    $\zeta$ & Entropy scaling parameter in the adaptive informative-cell threshold & 10 \\
    \bottomrule
  \end{tabularx}
\end{table*}



\subsection{Forward Gas Dispersion Model Evaluation}
\label{subsec:forward_model_evaluation}

\begin{figure*}
    \centering
    \includegraphics[width=0.85\linewidth]{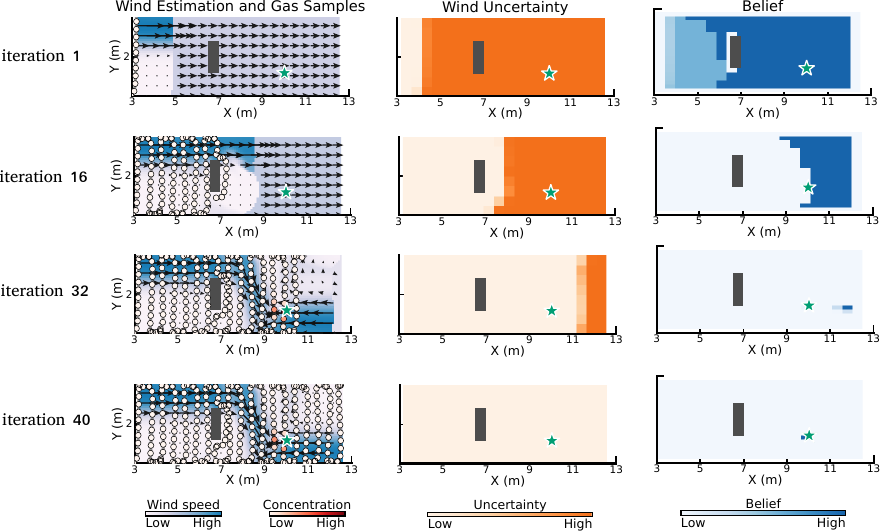}
    \caption {Evolution of the online wind reconstruction, wind-support uncertainty, and source belief during a representative lawnmower run in Map~4. Rows correspond to successive iterations; the red star marks the true source location.}
    \label{fig:online_wind_example}    
\end{figure*}

This experiment evaluates how the fidelity of the forward gas dispersion model affects source localization performance. To isolate this effect, all methods are evaluated using the same ground truth wind information and gas measurements collected along the identical lawnmower trajectory, as demonstrated in Fig.~\ref{fig:online_wind_example}. The localization process terminates when the robot reaches the end of the predefined path.
\begin{figure*}[t]
    \centering
    \includegraphics[width=0.8\linewidth]{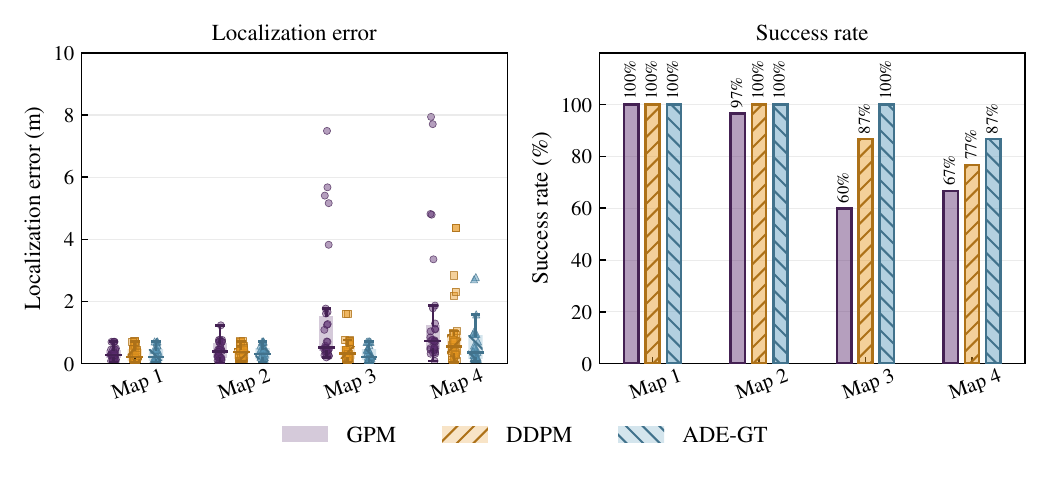}
    \caption{Localization performance of the evaluated forward dispersion models along the same predefined lawnmower trajectory. On the left side, the panels report localization error, and on the right side, success rate across the four simulated environments.}
    \label{fig:gas_models_evaluation}
\end{figure*}
The following forward models are compared:
\begin{itemize}
    \item \textbf{\ac{GPM}\cite{rahbar_design_2019}:}
    An analytical model which models the concentration at a point as a function of its distance to the source location, the source release rate, the average wind speed and standard deviations of gas dispersion in the Y-axis. It assumes approximately homogeneous airflow and does not explicitly represent obstacle-induced flow disturbances.
    

    \item \textbf{\ac{DDPM}\cite{jin_towards_2023}:}
    A U-Net based surrogate gas dispersion model trained with gas dispersion maps gathered in cluttered environments.
    It was trained using 200 wind-tunnel-like environments containing rectangular obstacles under fixed inlet, outlet, and wind speed configurations.

    \item \textbf{\ac{ADE}-GT:}
    The proposed \ac{ADE}-based forward model using the
    ground truth wind field. 
\end{itemize}

\begin{figure*}[t]
    \centering
    \includegraphics[width=0.8\linewidth]{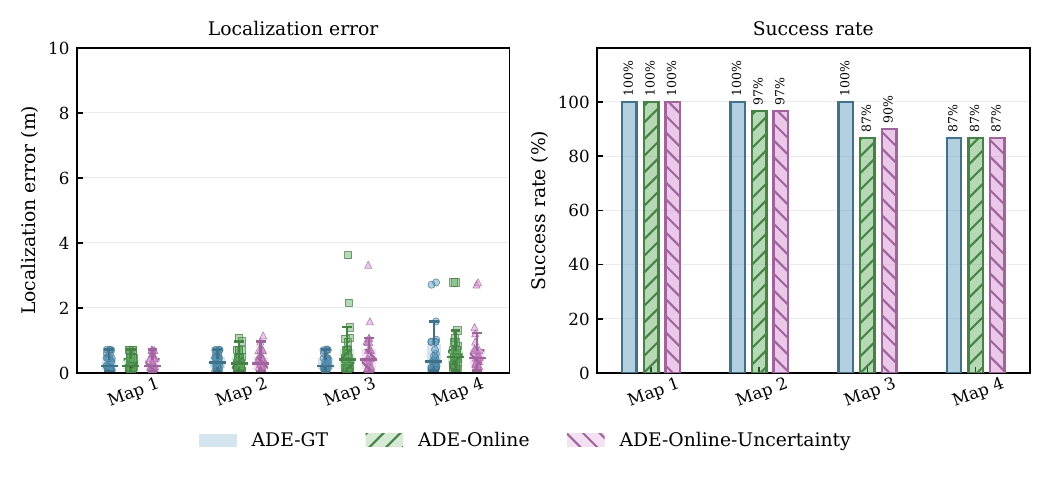}
    \caption{Effect of wind information and wind-support-aware inference on localization performance along the same predefined lawnmower trajectory. On the left side, the panels report localization error, and on the right side, success rate across the four simulated environments.}
    \label{fig:online_wind_comparison}
\end{figure*}

Figure~\ref{fig:gas_models_evaluation} compares the three forward models. In Map~1, all methods achieve similar localization performance because the airflow is approximately uniform and therefore agrees with the assumptions of the \ac{GPM}. All models still have decent performance in Map~2, as dense coverage along the lawnmower trajectory also provides sufficient observations to tolerate moderate model mismatch.

The performance differences become more pronounced as the airflow complexity increases. The \ac{GPM} degrades in Maps~3 and~4 because a single average wind vector cannot represent the spatially varying transport paths. The \ac{DDPM} performs well in environments similar to  its training dataset, such as Map~2 but degrades in Maps~3 and~4, where the obstacle interactions and multidirectional airflow differ from the training conditions. In contrast, \ac{ADE}-GT maintains more consistent performance because it directly incorporates the spatial wind field without relying on a fixed environmental configuration.

These results indicate that a numerical \ac{ADE} provides sufficient forward-model fidelity for source localization under spatially complex airflow, when a decent wind field estimate is available.

\subsection{Performance of Wind Estimation}
\label{subsec:online_wind_evaluation}

This experiment evaluates the effect of replacing the ground-truth wind field with measurements reconstructed online. All methods use the same lawnmower trajectory and therefore receive measurements at identical locations. We compare:

\begin{itemize}
    \item \textbf{\ac{ADE}-GT:}
    The \ac{ADE} supplied with the complete ground-truth wind field.

    \item \textbf{\ac{ADE}-Online:} Online wind reconstruction and \ac{ADE}-based source inference with $A_w=0$, resulting in the same baseline likelihood variance $\sigma_0^2$ for all source hypotheses.

    \item \textbf{\ac{ADE}-Online-Uncertainty:} The same pipeline with $A_w>0$, such that the likelihood variance additionally depends on the local wind support $W_k(E_i)$.
\end{itemize}

Figure~\ref{fig:online_wind_example} illustrates how the wind field and its support evolve during one run using the \ac{ADE}-Online-Uncertainty method. As the robot explores new regions, local kernel estimates replace the background wind approximation and the wind-support-aware uncertainty term $\lambda_k^w(E_i)$ decreases. The source belief is updated incrementally using the acquired gas and wind measurements. By the end of the exploration, the belief map converges near the true source location.

The results in Fig.~\ref{fig:online_wind_comparison} show that replacing the ground-truth wind field with online reconstruction has little effect in simple airflow scenarios. In these environments, measurements collected along the lawnmower path are sufficient to recover the dominant airflow direction. The performance gap increases in Maps~3 and~4, where local wake regions and multidirectional flow are more difficult to reconstruct from sparse measurements. Including the wind-support-aware uncertainty term provides a small improvement under the fixed trajectory. Because all methods visit the same locations, they eventually obtain similar spatial measurement coverage. The main benefit of \ac{ADE}-Online-Uncertainty is expected in closed-loop exploration, where early belief maps affect the subsequent trajectory.

\subsection{End-to-End Evaluation with IPP}
The previous experiments use a fixed trajectory to isolate the effects of the forward gas model and wind reconstruction from those of path planning. However, exhaustive lawnmower coverage can require long travel distances in large environments and cannot adapt to the evolving source belief. When the spacing between adjacent scan lines is large, the robot may fail to collect sufficiently informative measurements near the true source. In contrast, \ac{IPP} adapts the robot trajectory according to the current belief map and prioritizes regions that are more informative for source localization. This can reduce the required travel distance, but it also couples estimation and planning: an incorrect early belief may direct the robot toward uninformative regions and reinforce an incorrect source hypothesis.

We now evaluate the complete closed-loop system using the same belief-clustering \ac{IPP} strategy detailed in \ref{sec:IPP} for all methods. We compare the following combinations:

\begin{itemize}
    \item \ac{GPM} with \ac{IPP};
    \item \ac{DDPM} with \ac{IPP};
    \item \ac{ADE}-Online with \ac{IPP};
    \item \ac{ADE}-Online-Uncertainty with \ac{IPP}
\end{itemize}

\begin{figure*}
    \centering
    \includegraphics[width=0.85\linewidth]{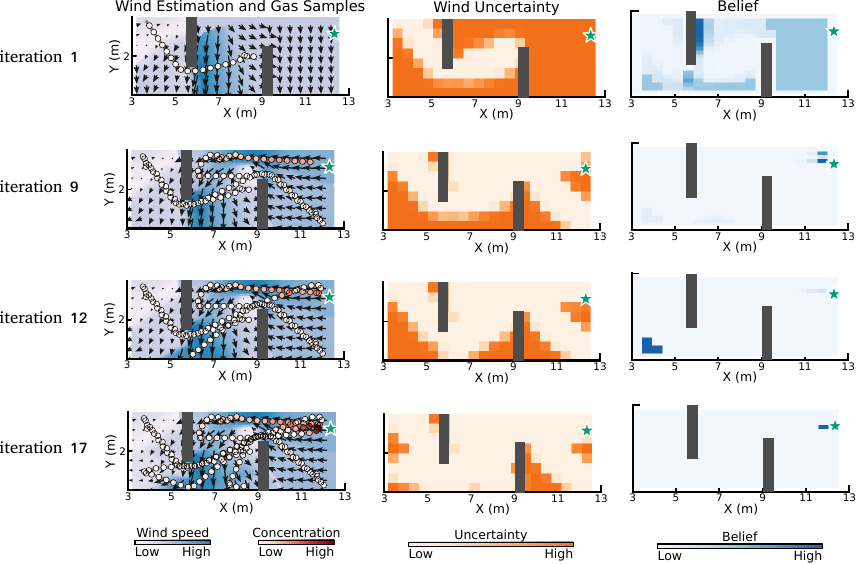}
    \caption{Evolution of a representative closed-loop \ac{GSL} run using \ac{ADE}-Online-Uncertainty with \ac{IPP} in Map~3. The panels show the collected gas measurements and reconstructed wind field, wind-support uncertainty, and source belief at successive iterations. The red star marks the true source location.}
    \label{fig:GSL_with_IPP_middle_results}  
\end{figure*}
\begin{figure*}
    \centering
    \includegraphics[width=\linewidth]{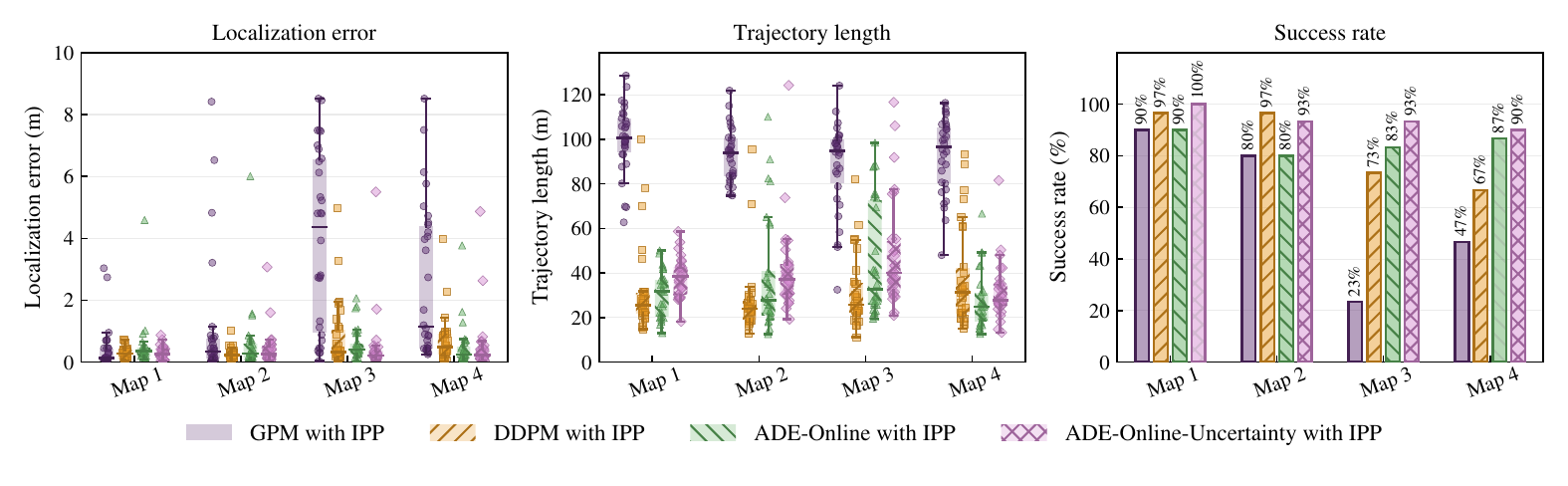}
    \caption{End-to-end localization performance of the evaluated dispersion models and wind-inference variants when combined with the same \ac{IPP} strategy. }
    \label{fig:entire_pipeline_evaluation}    
\end{figure*}

Figure~\ref{fig:GSL_with_IPP_middle_results} illustrates one run of the complete \ac{ADE}-Online-Uncertainty with \ac{IPP} framework. The robot follows the \ac{IPP} strategy by first scanning the informative region and then exploring the most informative region. Early in the search, the wind field near the true source is poorly supported and the corresponding additional mismatch tolerance remains high. The algorithm therefore avoids strongly penalizing that true source region based on an unreliable forward prediction. Consequently, the posterior probability associated with that region remains relatively high, allowing it to be selected for subsequent exploration. As new wind and gas measurements are collected, the wind reconstruction improves and the belief becomes concentrated near the true source.

The aggregate results for all the method in Fig.~\ref{fig:entire_pipeline_evaluation} show that the \ac{GPM} and \ac{DDPM} become increasingly sensitive to environmental complexity. Model mismatch can distort the belief, causing the planner to repeatedly visit incorrect regions. The \ac{ADE}-Online provides more consistent performance by incorporating the spatially reconstructed wind field. Adding wind-support-aware uncertainty further reduces the risk of premature convergence during the early stages of exploration, when the reconstructed wind field is still incomplete. On the other hand, this more conservative treatment may increase travel distance in runs where the source region is identified early, because poorly explored alternatives are retained for longer.

Overall, \ac{ADE}-Online-Uncertainty achieves the most consistent success rate and localization performance across the evaluated scenarios. Compared with the predefined lawnmower trajectory, integrating \ac{ADE}-Online-Uncertainty with \ac{IPP} substantially reduces travel distance by more than half while maintaining or improving the success rate in the more complex environments.


\section{Physical Experiments} \label{sec:physical_exp}

\begin{figure}
    \centering
    \includegraphics[width=\linewidth]{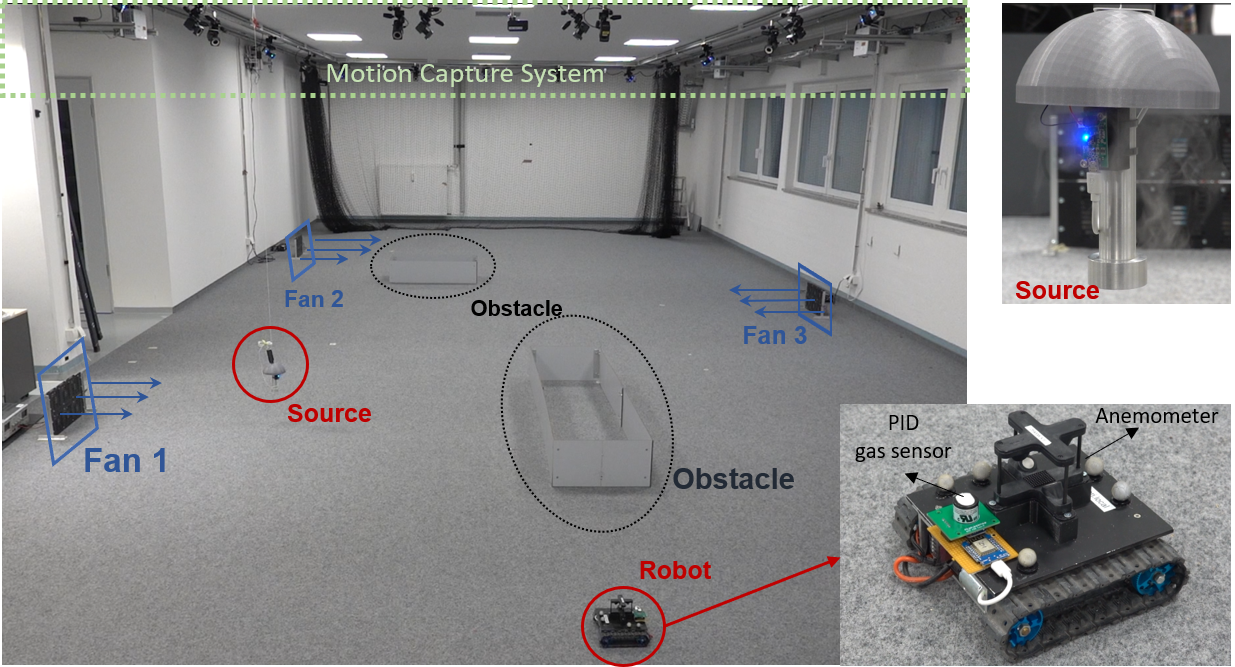}
    \caption{Physical experimental platform and indoor test environment. Multiple fans generate spatially varying airflow, while the mobile robot carries an anemometer and a gas sensor for continuous wind and gas measurement acquisition.}
    \label{fig:experimental_setup}
\end{figure}

\begin{figure}
    \centering
    \includegraphics[width=0.7\linewidth]{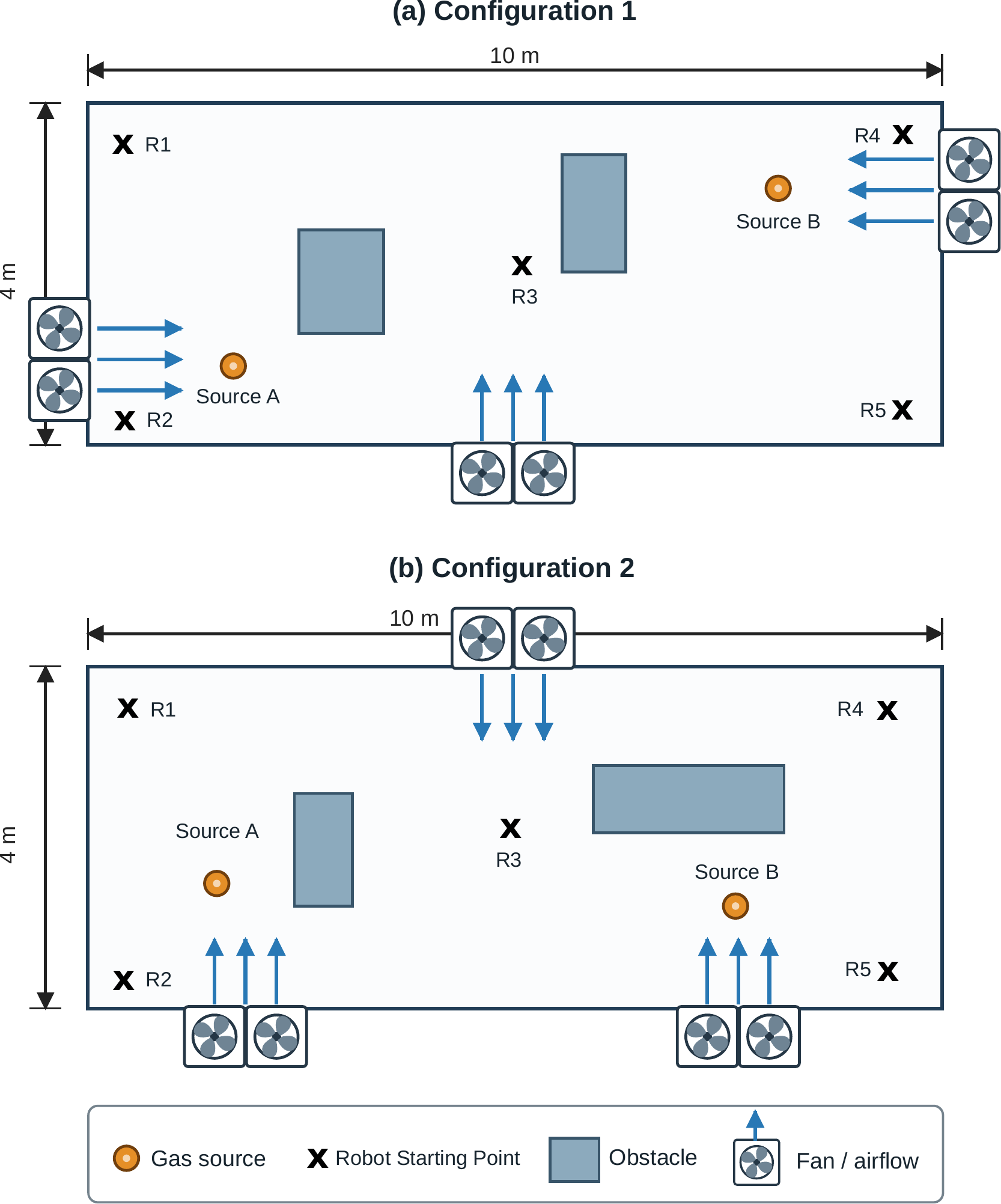}
    \caption{Layouts of the two physical experimental setups. Each setup contains two different source configurations, denoted A and B, together with obstacles, fan locations, and the candidate robot starting positions.}
    \label{fig:experimental_config}
\end{figure}

The physical experiments evaluate the complete \ac{GSL} framework under real sensor performance, airflow disturbances, and gas dispersion phenomena. 

\subsection{Experimental Setup}
\label{subsec:physical_setup}

\subsubsection{Environment and airflow conditions}

The experiments are conducted in a $13\,\mathrm{m}\times4\,\mathrm{m}$ indoor arena, within which a $10\,\mathrm{m}\times4\,\mathrm{m}$ region is used as the robot's search workspace. We construct two testing configurations containing rectangular obstacles of different sizes and at different locations, as shown in Fig.~\ref{fig:experimental_config}. In addition to the background airflow generated by the ventilation and air conditioning systems, three fan arrays are placed around the arena and oriented in different directions. The resulting airflow exhibits substantial spatial variation and resembles the multidirectional conditions represented by Map~4 in simulation.

The physical inlet and outlet conditions are not provided to the \ac{GSL} algorithm. Accordingly, the \ac{ADE} solver uses the default boundary conditions described in Sec.~\ref{sec:boundary_condition}. Dirichlet boundary conditions are applied at the outer boundary of the experimental domain, and no explicit obstacle boundary conditions are imposed. The reconstructed wind field, therefore, provides the primary representation of obstacle-induced airflow effects. The obstacle map used by the navigation system is not used to prescribe physical airflow boundary conditions in the dispersion model.

\subsubsection{Gas source and robotic platform}
The gas source and the robot used in the physical experiments are shown in Fig.~\ref{fig:experimental_setup}. The gas source is emulated using a vaporizer filled with a mixture of ethanol and water, therefore, the source release rate is not controllable. A custom outlet hood is mounted above the vaporizer to direct the released vapor downwards. The source is fixed on the ceiling slightly above the robot height, allowing the robot to pass underneath without collision. The source locations are positioned in front of the fan arrays to promote gas dispersion and help the plume approach a steady time-averaged state. The arena was ventilated between runs by opening the windows.

The experiments use a non-holonomic rover. Two tracks allow the robot to turn and move forward and backward. The robot is equipped with a TriSonica Mini anemometer \cite{trisonica_website} and an Alphasense PID-AH5 gas sensor \cite{pid_website}. The anemometer provides planar wind measurements in a range from 0–50 $\mathrm{m/s}$ with an accuracy of $±0.2$ $\mathrm{m/s}$ and a resolution of 0.01 $\mathrm{m/s}$, while the \ac{PID} provides the raw gas sensor signal, reported in voltage units and positively related to the local gas concentration. A motion capture system supplies the robot pose required for navigation and associates each sensor measurement with its sampling location.

Wind reconstruction, \ac{ADE} prediction, Bayesian source inference, and \ac{IPP} are executed on an Intel Core i7-12700H computer. Sensor measurements and robot locations are transmitted to the computer, which returns navigation waypoints to the robot. The maximum robot speed is set to 0.3 $\mathrm{m/s}$, and wind and gas measurements are sampled at 10 Hz.

\subsubsection{Experimental protocol}
The physical experiments compare \ac{ADE}-Online and
\ac{ADE}-Online-Uncertainty, using the same definitions and parameter settings
as in the simulation experiments.

Two source locations are evaluated in each environmental configuration (marked as Source A/B in Fig.~\ref{fig:experimental_config}). For each source location, the robot is initialized from five positions: the four corners of the search region and its center (marked as R 1-5 in Fig.~\ref{fig:experimental_config}). Each combination of environment, source location, starting position, and localization method is repeated three times. Each method is therefore evaluated over $2\times2\times5\times3=60$ runs, resulting in 120 physical runs in total. Localization error and travel distance are evaluated using the definitions introduced in Sec.~\ref{subsec: simulation_setup}.

\subsection{Physical Experiment Results}
\label{subsec:physical_results}
\begin{figure}
    \centering
    \includegraphics[width=\linewidth]{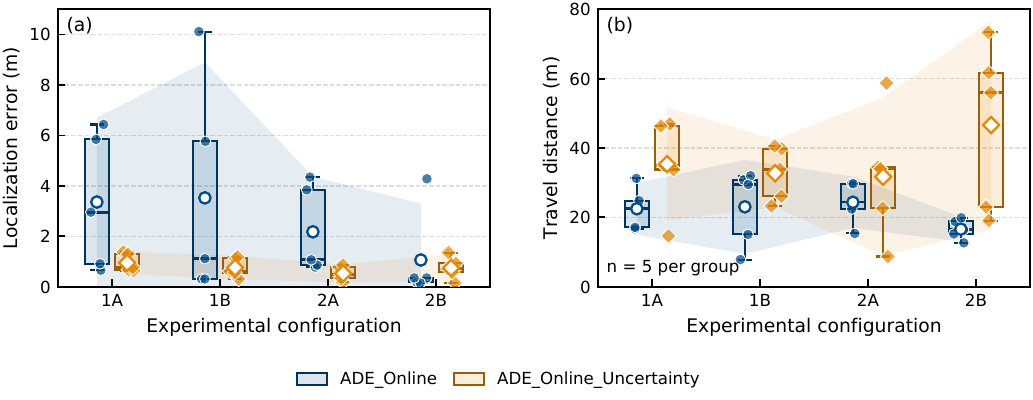}
    \caption{Physical experimental results for \ac{ADE}-Online and \ac{ADE}-Online-Uncertainty across the four source configurations. The panels report the localization-error and travel-distance distributions over repeated trials.}
    \label{fig:physical_experiment_result}
\end{figure}

\begin{table}[t]
    \centering
    \caption{Physical experimental performance summary}
    \label{tab:physical_results}
    \begin{tabular}{lccc}
        \hline
        \textbf{Method} &
        \textbf{Error[m]} &
        \textbf{Traveling } &
        \textbf{Duration [s]} \\
         &
         &
        \textbf{Distance [m] } &
         \\
        \hline
        \ac{ADE}-Online &
        2.5 ± 2.8 &
        21.6 ± 7.1 &
        166.2 ± 42.9 \\
        \ac{ADE}-Online-Uncertainty &
        0.7 ± 0.3 &
        36.6 ± 16.7 &
        219.5 ± 88.9 \\
        \hline
    \end{tabular}
\end{table}

Table~\ref{tab:physical_results} provides a quantitative summary of the localization performance, while Fig.~\ref{fig:physical_experiment_result} shows the corresponding distributions across experimental trials. Overall, \textit{\ac{ADE}-Online-Uncertainty} achieves lower localization errors and exhibits a smaller variance in localization error than \textit{\ac{ADE}-Online} across the tested configurations. The improvement is most pronounced in Configurations 1A, 1B, and 2A. In these configurations, the airflow near the source differs from the dominant airflow in the rest of the environment, and the source vicinity is not sufficiently sampled during the early stages of exploration.

\begin{figure}[t]
    \centering

    \begin{minipage}{\linewidth}
        \centering
        \includegraphics[width=\linewidth]{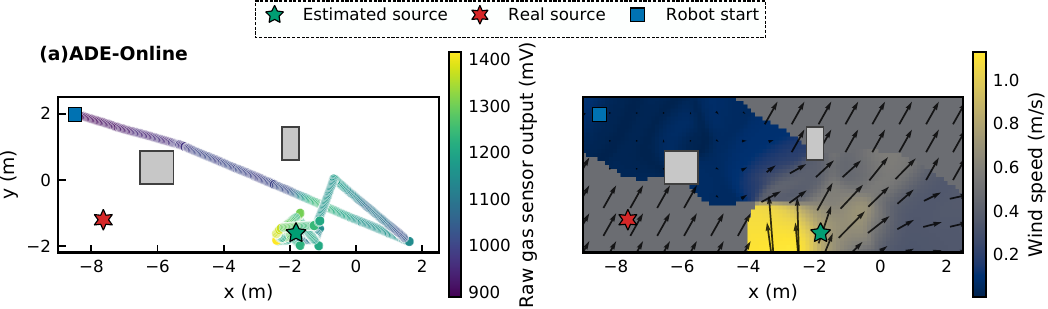}
    \end{minipage}

    \begin{minipage}{\linewidth}
        \centering
        \includegraphics[width=\linewidth]{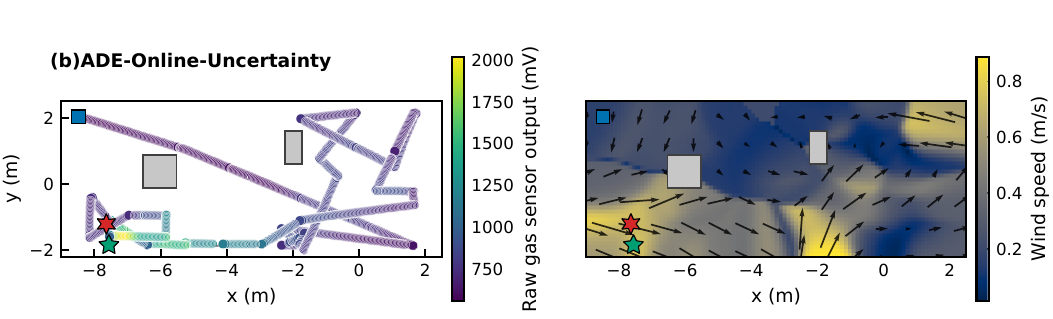}
    \end{minipage}

    \caption{Example of two runs with ADE-Online and ADE-Online-Uncertainty in Config 1A. Left: robot trajectory and the raw gas measurements. Right: the reconstructed wind field.}
    \label{fig:physcial_exp_individual_run}
\end{figure}

Without the wind-support-aware formulation, discrepancies caused by the initially incomplete wind reconstruction can strongly suppress source hypotheses in poorly explored regions. Because the \ac{IPP} strategy is guided by the resulting belief map, an incorrect early update may direct the robot toward an incorrect region and reduce the likelihood of subsequently exploring the vicinity of the true source. Consequently, \textit{\ac{ADE}-Online} prematurely converges to an incorrect source hypothesis in several trials. One representative failure run is shown in Fig.\ref{fig:physcial_exp_individual_run} (a).

In contrast, \textit{\ac{ADE}-Online-Uncertainty} assigns greater mismatch tolerance to source hypotheses located in regions with limited wind measurement support. These hypotheses therefore retain plausibility of containing the source until additional wind measurements are collected nearby. As the local support increases, the additional tolerance decreases, allowing agreement between the measured and predicted gas patterns to exert a stronger influence on the posterior. This more conservative treatment reduces the risk of prematurely eliminating the true source hypothesis when the reconstructed wind field is still incomplete, therefore improving robustness to variations in the robot's initial position and source location across the tested configurations. One representative successful run is shown in Fig.\ref{fig:physcial_exp_individual_run} (b).

This conservativeness can, however, increase the required trajectory length when the correct source region is encountered early, as observed in Configuration 2B. In this configuration, the source is located closer to the center of the environment and is therefore more likely to be approached regardless of the robot's initial position. Although the measurements may support the correct source region at an early stage, alternative hypotheses in underexplored regions are retained for longer while using \textit{\ac{ADE}-Online-Uncertainty}, delaying posterior concentration and resulting in slower termination.






\section{Conclusion} \label{sec:conclusion}

This paper presented a lightweight physics-informed framework for online gas source localization under spatially complex indoor airflow. The framework reconstructs the wind field incrementally from measurements collected by a mobile robot and uses the resulting field in an \ac{ADE}-based forward model to predict the relative gas distribution pattern associated with each candidate source location. Unlike approaches that rely on a precomputed air flow field or detailed environmental boundary conditions, the proposed method can operate without prior knowledge of obstacle geometry or inlet and outlet locations. Furthermore, a wind-support-aware likelihood formulation reduces the influence of predictions in regions where the reconstructed wind field is poorly supported, mitigating premature elimination of the true source hypothesis.

The simulation experiments showed that incorporating spatially varying airflow through the ADE model provides more consistent localization performance than simplified analytical plume models and \ac{DDPM}, particularly in environments containing obstacles and multidirectional airflow. The ablation studies further demonstrated the respective effects of online wind reconstruction and wind-support-aware inference. When integrated with informative path planning, the proposed framework maintained robust localization performance while substantially reducing the required travel distance compared with exhaustive predefined trajectories. The physical experiments further demonstrated the feasibility of continuously collecting wind and gas measurements during robot motion and executing the complete framework online under unknown indoor airflow conditions. Overall, the results suggest that a computationally lightweight, task-oriented dispersion model can provide sufficient physical guidance for efficient GSL without requiring a high-fidelity reconstruction of the complete flow and concentration fields.

The present work nevertheless has several limitations. First, it assumes a single stationary source and approximately steady time-averaged airflow and gas distribution fields during each localization task. Extending the framework to time varying airflow, intermittent releases, moving sources, and multiple sources remains an important direction for future work. Second, localization becomes more challenging when the source lies in a region with very low wind speed, where the measured airflow provides limited information about the dominant gas transport direction. Future work could incorporate additional diffusion information or temporal gas features to improve inference under weak advection conditions. Finally, the proposed wind-support term evaluates the measurement support near each candidate source and serves as a lightweight proxy for the reliability of its forward prediction. However, it does not explicitly propagate wind field uncertainty through the complete downstream plume. A more principled uncertainty model that accounts for wind uncertainty along the entire transport path and propagates it through the \ac{ADE} solver could provide a more accurate source-dependent likelihood while retaining online computational efficiency.







\bibliographystyle{IEEEtran}
\bibliography{IEEEexample}

\end{document}